\documentclass{arxivpaper}

\usepackage{xspace}
\usepackage[normalem]{ulem}
\usepackage[export]{adjustbox}
\usepackage{enumitem}

\usepackage{tabularx}

\graphicspath{{figs/}}

\makeatletter
\DeclareRobustCommand\onedot{\futurelet\@let@token\@onedot}
\def\@onedot{\ifx\@let@token.\else.\null\fi\xspace}

\makeatother

\newtcolorbox{graylist}{
  enhanced,
  colback=gray!5,
  colframe=gray!15,
  boxrule=0pt,
  arc=0mm,
  left=1mm, right=1mm, top=1mm, bottom=1mm,
  width=\dimexpr\linewidth-4mm\relax,
  center,
  breakable
}

\useunder{\uline}{\ul}{}

\newcommand{\name}{\textnormal{\textsc{ChaosMining}}\xspace}
\newcommand{\projectrepo}[1]{\href{https://github.com/geshijoker/ChaosMining/}{#1}}
\newcommand{\datasetrepo}[1]{\href{https://huggingface.co/datasets/geshijoker/chaosmining}{#1}}
\newcommand{\datasetdiscussion}[1]{\href{https://huggingface.co/datasets/geshijoker/chaosmining/discussions}{#1}}
\newcommand{\datasetdoi}[1]{#1}
\newcommand{\appendixlocation}[1]{Appendix~\ref{#1}}
\newcommand{\mainpageextension}{}
\newenvironment{audiotable}{\begin{table*}[!htbp]}{\end{table*}}

\setlist[itemize]{leftmargin=*}

\let\cite\citep

\title{\name: Benchmarking Post-Hoc Attribution with Sparse Informative Features in High Dimensions}
\abstract{Post-hoc attribution is widely used to identify important model inputs, but
evaluating whether these attributions identify truly informative features is
difficult because real datasets rarely provide reliable ground truth. We
introduce a multimodal benchmark containing symbolic tabular, vision, and audio
tasks with known informative feature sets. In the main benchmark conditions,
informative variables, spatial regions, or channels occupy fixed input
coordinates while the remaining inputs provide irrelevant or distracting
information. We use the benchmark to study how attribution quality depends on
predictive performance, irrelevant-feature burden and structure, model
configuration, and attribution mechanism, while separately measuring
identification, stability, and computational cost. In most symbolic-data
sweeps, informative-set identification co-varies with predictive performance,
while the relative ordering of attribution methods remains largely stable. Across
modalities, no method dominates all architectures and conditions, and greater
attribution complexity does not consistently improve identification. Simple
gradient attribution is often competitive at lower computational cost, while
the vision and audio results show that architecture and the form of irrelevant
content materially affect attribution quality.
}

\author[1]{Ge Shi}
\author[2]{Fangyi Liu}
\author[1]{Ziwen Kan}
\author[3]{Menglin Liu}

\affiliation[1]{University of California, Davis, United States}
\affiliation[2]{Nankai University, China}
\affiliation[3]{The Chinese University of Hong Kong, Shenzhen, China}

\metadata[Code]{\url{https://github.com/geshijoker/ChaosMining/}}
\metadata[Data]{\url{https://huggingface.co/datasets/geshijoker/chaosmining}}
\correspondence{Menglin Liu (\texttt{menglinliu@cuhk.edu.cn})}

\begin{document}

\maketitle

\section{Introduction}\label{sec:1}

High-dimensional inputs often contain only a small set of variables, spatial
locations, or channels that are informative for prediction. The remaining
inputs may be statistically independent noise, structured but task-irrelevant
content, or correlated alternatives to the informative variables. Identifying
the informative inputs is important for model interpretation, scientific
analysis, and dimensionality reduction, particularly in applications where
irrelevant features are common \cite{caiafa2021machine,roscher2020explainable}.

Post-hoc local attribution methods assign an importance value to each input of
an already-trained model \cite{ancona2017towards,zhang2021attention,
zhang2022improving}. These scores are increasingly used not only to explain
individual predictions, but also to rank and select features in genomics,
energy, agriculture, cybersecurity, software engineering, and drug discovery
\cite{al-absii2023xai,yu2026hi4gs,alotaibi2024xai,
bhatia2025xai,vamathevan2024xai}. Evaluating this use requires distinguishing
inputs emphasized by a trained predictor from inputs that truly generate the
target. A predictor can rely on irrelevant, redundant, or spuriously correlated
features, so a faithful explanation of that predictor need not identify the
ground-truth informative set.

Direct evaluation is difficult on observational data. The real tabular
datasets commonly used for classification and regression provide downstream
performance and stability measurements, but generally do not reveal which
features truly generated the target. Existing controlled XAI benchmarks offer
valuable synthetic functions, annotated image regions, or domain-specific
tests \cite{liu2021synthetic,agarwal2024openxai,arras2022clevrxai}; however,
ground-truth evaluations remain fragmented across data types and objectives.
There is limited evidence about whether conclusions obtained from tabular
variables extend to spatial regions or sequential channels, or how attribution
changes when a small informative set is surrounded by many irrelevant inputs.

We introduce \name, a benchmark with known informative feature sets in three
modalities. Symbolic regression tasks provide human-defined nonlinear functions
and exact informative variables. Vision tasks embed a class-bearing foreground
inside a larger irrelevant background. Audio tasks place a predictive speech
signal in one channel among irrelevant channels. The main conditions use fixed
feature correspondence across samples: the same variable identities, spatial
region, or channel positions are informative throughout a dataset. This design
matches the consistent feature semantics of tabular data while extending the
evaluation to structured inputs.
Accordingly, \name evaluates whether post-hoc attributions recover ground-truth
informative features, not whether they are faithful to a model's decision
process.

\begin{figure*}[htbp]
    \centering
    \begin{subfigure}[b]{0.31\linewidth}
        \centering
        \includegraphics[width=\linewidth]{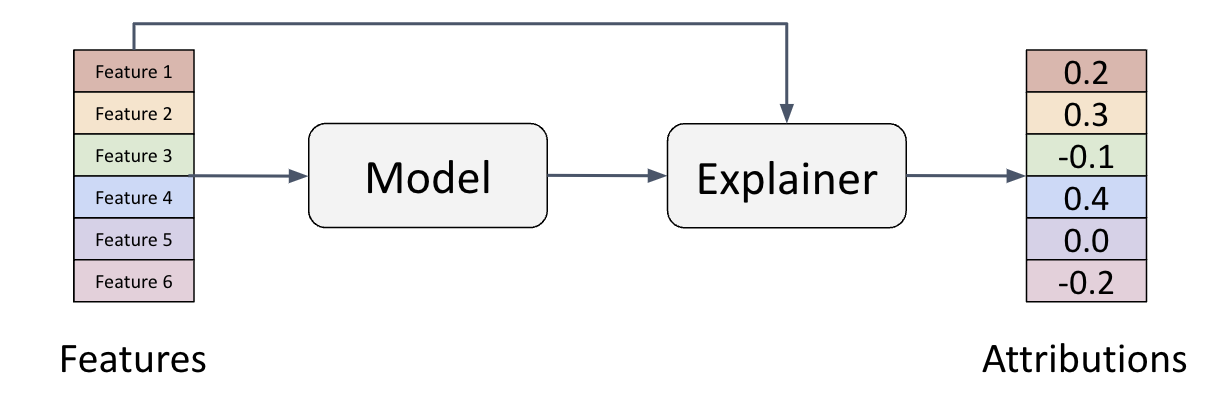}
        \caption{Local feature attribution.}
        \label{fig:sub1}
    \end{subfigure}
    \hfill
    \begin{subfigure}[b]{0.31\linewidth}
        \centering
        \includegraphics[width=\linewidth]{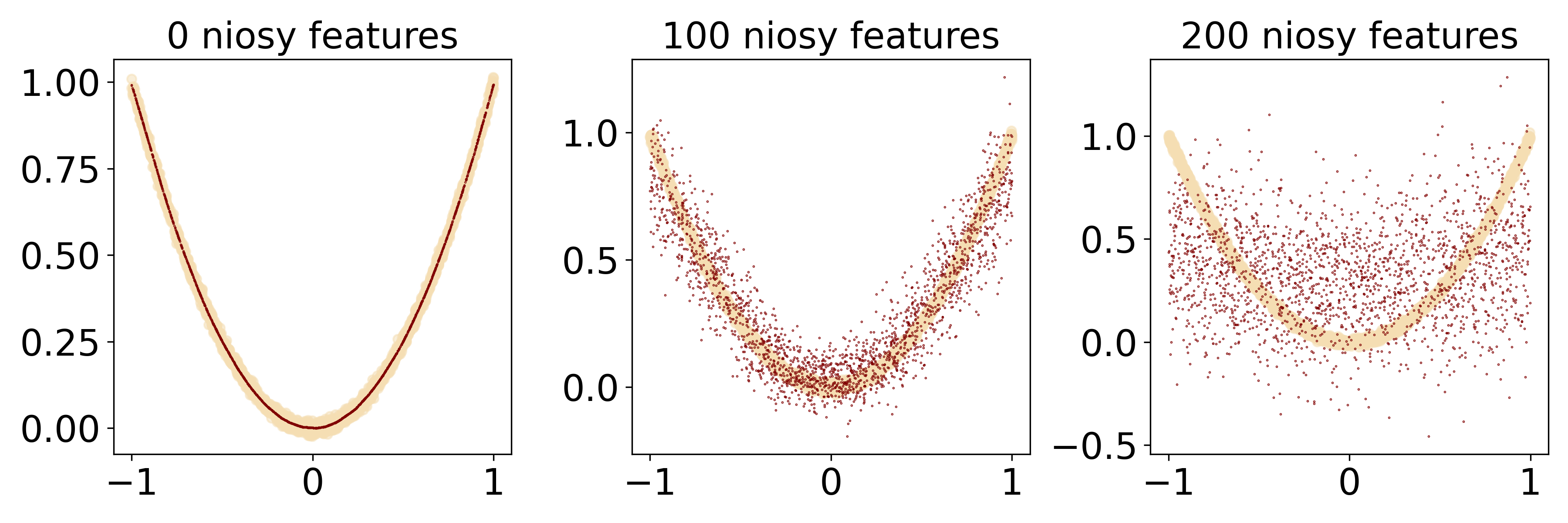}
        \caption{Irrelevant-feature burden.}
        \label{fig:sub2}
    \end{subfigure}
    \hfill
    \begin{subfigure}[b]{0.31\linewidth}
        \centering
        \includegraphics[width=\linewidth]{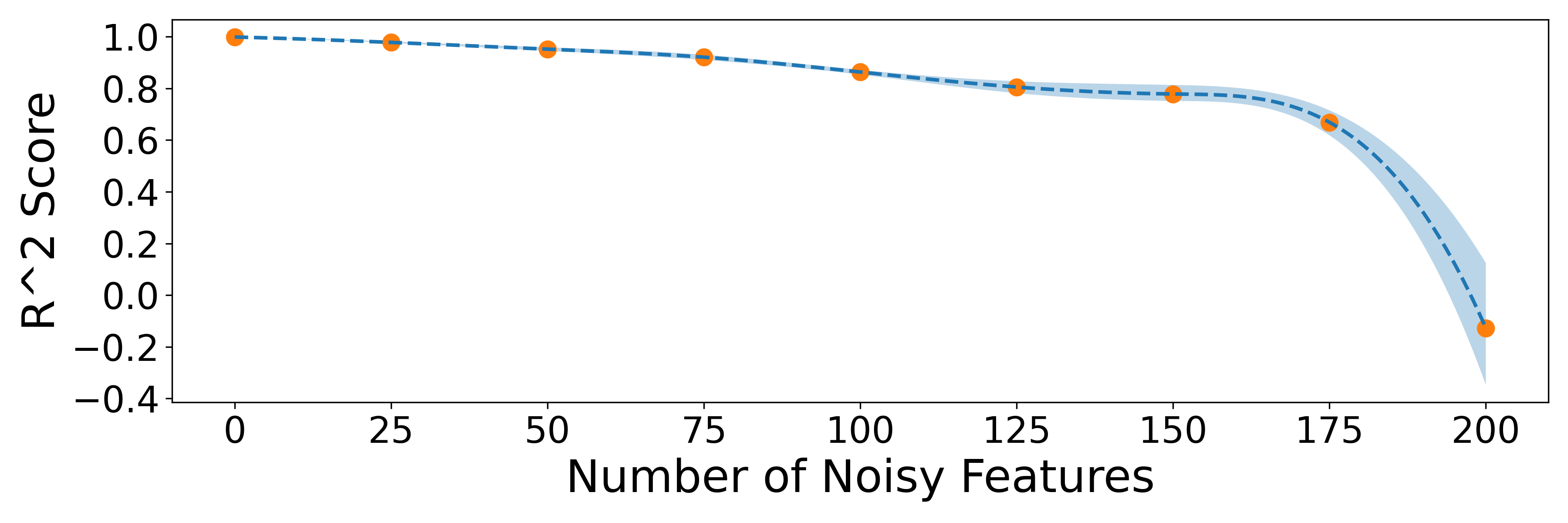}
        \caption{Prediction under noise.}
        \label{fig:sub3}
    \end{subfigure}
    \caption{Benchmark motivation. Attribution scores are evaluated against a
    known informative feature set while irrelevant inputs and model conditions
    are varied.}
    \label{fig:complete}
\end{figure*}

The study addresses the following questions. First, how do irrelevant-feature
burden, data availability, label noise, and predictive performance affect the
identification of informative features? Second, does a more elaborate or
computationally expensive attribution procedure consistently provide a better
ranking, greater stability, or better computational efficiency? Third, do
patterns observed for tabular variables persist for spatial and
channel-structured inputs?

Across the experiments, attribution quality deteriorates when the underlying
predictor fails to generalize, but strong prediction alone does not establish
correct feature identification. No attribution method dominates all models and
data conditions. Simple gradient Saliency is frequently competitive and has low
computational overhead. The structured-modality experiments further show that
model architecture and the form of irrelevant content materially affect
attribution.

Our contributions are three-fold:

\begin{itemize}[leftmargin=*]
    \item \textbf{Benchmark construction.} We introduce a multimodal benchmark
    with known informative feature sets and controllable irrelevant inputs
    across symbolic tabular, vision, and audio data.
    \item \textbf{Systematic empirical study.} We evaluate how informative-
    feature identification depends on predictive performance, input and model
    conditions, attribution stability, and computational cost.
    \item \textbf{Cross-modal characterization.} We identify findings that
    persist across symbolic variables, image regions, and audio channels, while
    documenting architecture- and modality-specific differences that prevent a
    universal attribution-method ranking.
\end{itemize}

Section~\ref{sec:related} reviews attribution methods and controlled evaluation.
Section~\ref{sec:pip} formalizes the benchmark and metrics.
Section~\ref{sec:exp} presents the controlled experiments, and
Section~\ref{sec:discussion} synthesizes the findings and limitations.

\section{Related Work}\label{sec:related}

\paragraph{Post-hoc attribution.}
Feature-attribution methods assign importance values to the inputs of a trained
predictor. The methods evaluated in this work span local gradients
\cite{simonyan2013deep}, path integration \cite{sundararajan2017axiomatic},
activation-difference propagation \cite{shrikumar2017learning}, perturbation
\cite{zeiler2014visualizing}, local surrogate models \cite{ribeiro2016why},
Shapley-value approximations \cite{lundberg2017unified}, guided gradients
\cite{springenberg2015striving}. These approaches differ in their computational
cost and in the model behavior that their scores approximate; greater
methodological complexity therefore does not by itself imply better
identification of the data-generating features.

\paragraph{Evaluation with ground truth.}
Attribution is commonly evaluated through faithfulness, sensitivity,
randomization, or perturbation tests \cite{JMLR:v24:22-0142,
belaid2023compare}. Controlled datasets instead permit direct comparison with
known informative variables or masks. Synthetic tabular benchmarks
\cite{liu2021synthetic,agarwal2024openxai} and studies of suppressor variables
\cite{wilming2022scrutinizing} provide known feature relationships, while
CLEVR-XAI supplies spatial ground-truth masks for vision explanations
\cite{arras2022clevrxai}. Other benchmarks emphasize graphs, language,
vision--language systems, or cross-model faithfulness
\cite{rathee2022bagel,wang2022fine,kim2022hive,li2023m4}. \name complements
these resources by using a common informative-set perspective across symbolic
variables, image regions, and audio channels while explicitly varying the
burden and form of irrelevant inputs.

\section{Benchmark Definition}\label{sec:pip}

\subsection{Ground-Truth Informative Sets}

Let \(x\in\mathbb{R}^{d}\) denote an input and let \(S^{\star}\) denote the
coordinates on which the target-generating process depends. We refer to
\(S^{\star}\) as the \emph{ground-truth informative set}. It specifies feature
membership, not an exact attribution magnitude: a variable can belong to
\(S^{\star}\) even when its local contribution is zero for a particular sample.
The benchmark asks whether a local attribution ranking concentrates on
\(S^{\star}\), rather than whether it reproduces a unique causal explanation.

The main benchmark uses \emph{fixed feature correspondence}: informative
variables, image regions, or audio channels occupy the same input coordinates
across samples. This is the natural setting for tabular variables and permits
local attributions to be aggregated without first solving a registration or
permutation problem. Random-position vision conditions are retained as a
secondary analysis of sensitivity to moving spatial support.

We describe benchmark difficulty through the number and form of irrelevant
inputs rather than through a power-based signal-to-noise ratio. For each
modality, an informative proportion can be reported as

\begin{equation}
    r_{\mathrm{info}}=\frac{|S^{\star}|}{d},
    \label{eq:info-ratio}
\end{equation}

where coordinates represent variables, evaluated image units, or channels as
appropriate. These modality-specific proportions characterize within-modality
sparsity and are not assumed to be numerically comparable across modalities.

\subsection{Benchmark Components}

Each experiment is organized as a
\emph{data condition \(\times\) predictive model \(\times\) attribution method}
combination. A model is a trained checkpoint determined by its architecture and
optimization configuration. Attribution methods are applied after training;
where methods share a reported condition, they use the same checkpoint.

The evaluated attribution procedures are Saliency (SA), DeepLIFT (DL),
Integrated Gradients (IG), Feature Ablation (FA), Kernel SHAP (KS), LIME,
DeepLIFT SHAP (DLS), and Guided Backpropagation (GB). Method
availability depends on the modality and architecture, and unavailable
combinations are reported explicitly rather than imputed.

\paragraph{Method families.}
For a predictor \(F\), input \(x\), and optional reference \(\bar{x}\), each
method returns a coordinate-wise attribution vector \(A_F(x,\bar{x})\).
SA and GB modify input gradients; IG integrates gradients along a reference
path; DL and DLS propagate reference-relative contributions; FA perturbs input
coordinates; and KS and LIME fit local surrogate models. We use the standard
formulations implemented by Captum.

\paragraph{Attribution implementation.}
We use the implementations provided by Captum~\cite{kokhlikyan2020captum}
rather than reimplementing the algorithms. Table~\ref{tab:attribution-settings}
reports the settings used in the experiments. ``Mean feature'' denotes the
vector of feature-wise means used as the reference input. Within each condition,
all applicable methods explain the same trained checkpoint and scalar model
output. Absolute attribution magnitudes are evaluated at the native benchmark
unit---input variables for symbolic data, spatial units for vision, and channels
after temporal aggregation for audio.

\begin{table}[!t]
  \centering
  \caption{Captum implementations and attribution parameters.}
  \label{tab:attribution-settings}
  \footnotesize
  \setlength{\tabcolsep}{3pt}
  \begin{tabularx}{\columnwidth}{@{}lX@{}}
    \toprule
    Method & Captum implementation and settings \\
    \midrule
    SA & \texttt{Saliency}; \texttt{abs=True} \\
    IG & \texttt{IntegratedGradients}; 10 steps; mean-feature baseline \\
    DL & \texttt{DeepLift}; mean-feature baseline \\
    DLS & \texttt{DeepLiftShap}; 50 baseline samples \\
    FA & \texttt{FeatureAblation}; mean-feature baseline \\
    GB & \texttt{GuidedBackprop}; Captum defaults \\
    LIME & \texttt{Lime}; 300 samples; mean-feature baseline \\
    KS & \texttt{KernelShap}; 300 samples; mean-feature baseline \\
    \bottomrule
  \end{tabularx}
\end{table}

\subsection{Data Generation}

\begin{figure}[ht]
\centering
\begin{subfigure}[b]{0.24\textwidth}
    \centering
    \includegraphics[width=\linewidth]{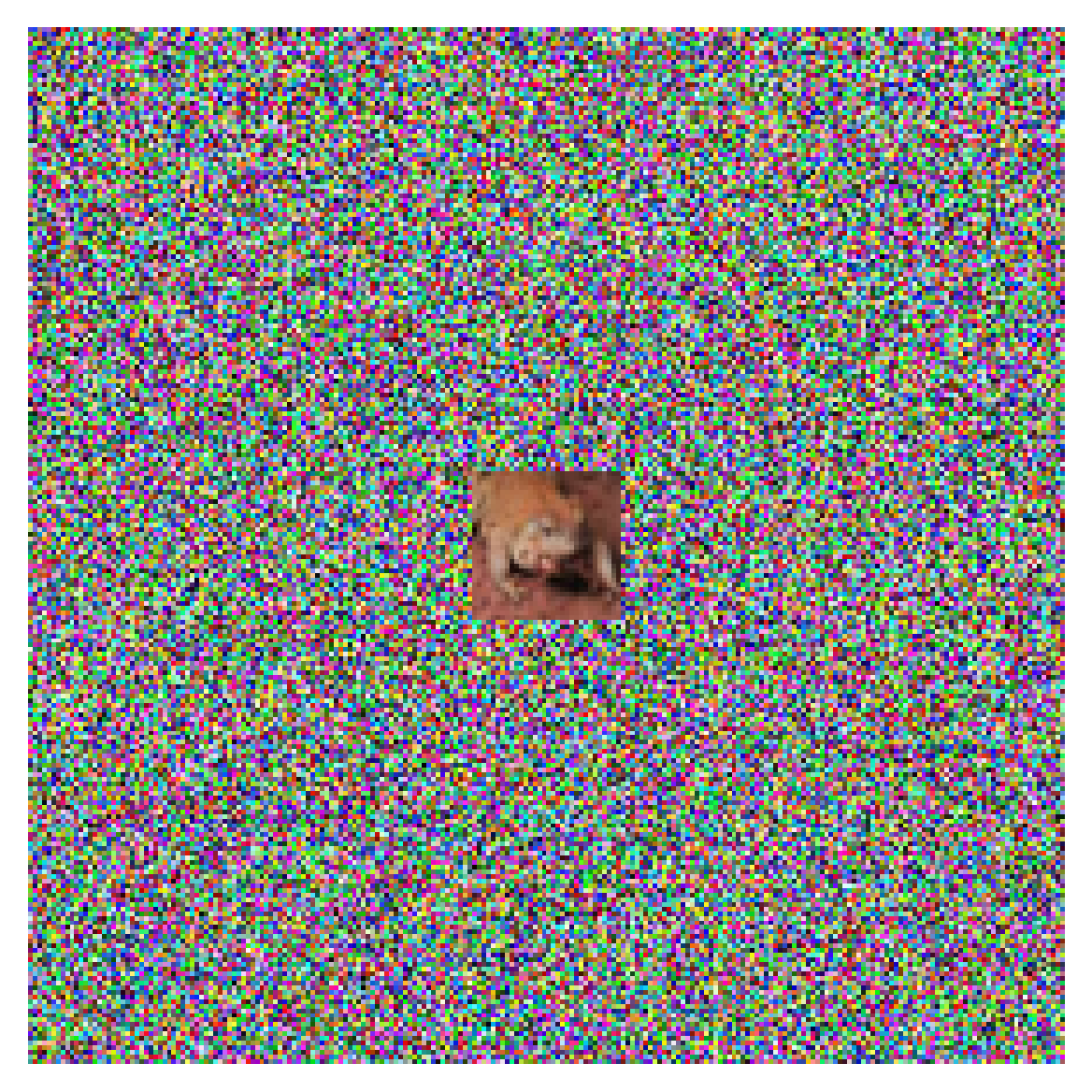}
    \caption{RBFP}
    \label{fig:rbfp}
\end{subfigure}
\hfill
\begin{subfigure}[b]{0.24\textwidth}
    \centering
    \includegraphics[width=\linewidth]{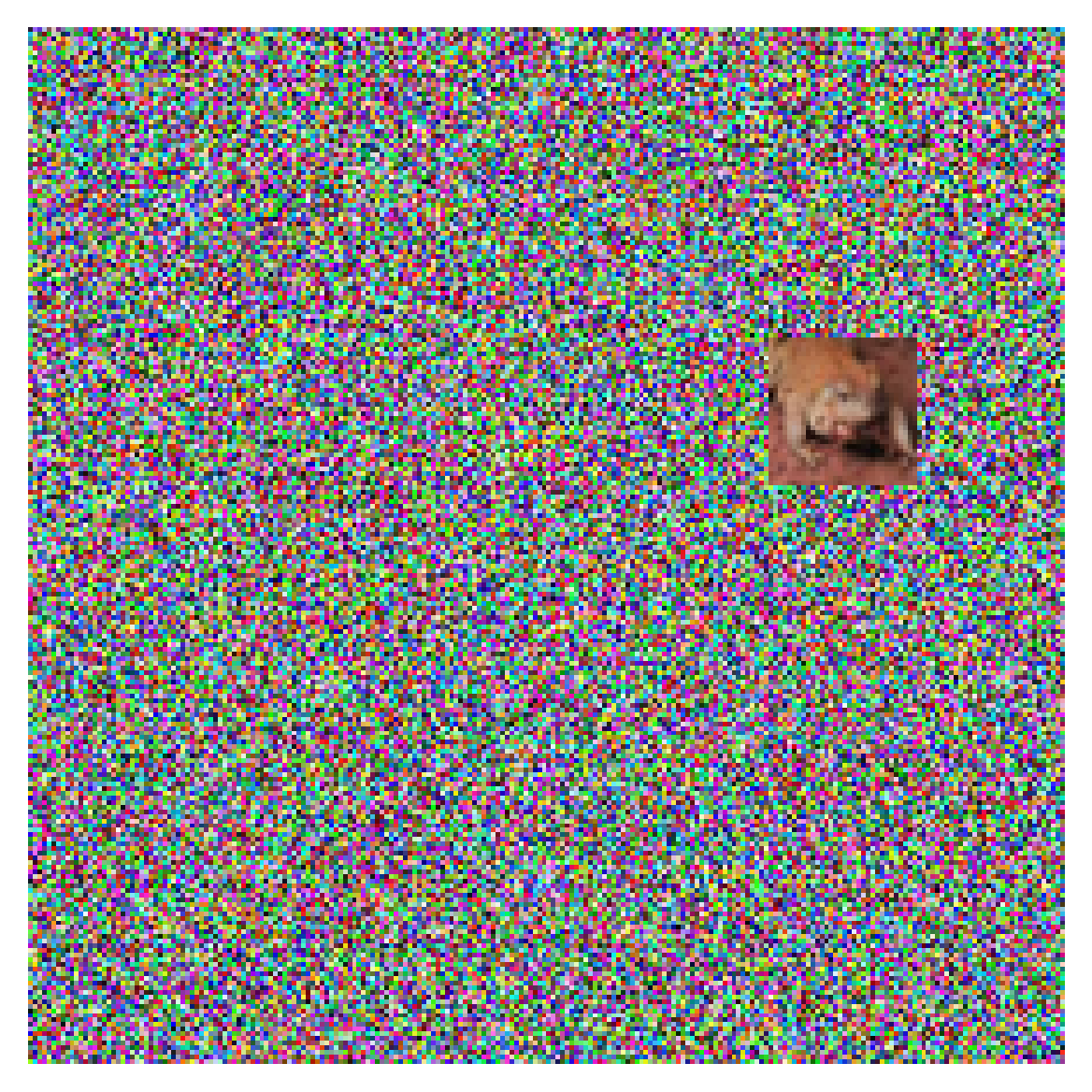}
    \caption{RBRP}
    \label{fig:rbrp}
\end{subfigure}
\hfill
\begin{subfigure}[b]{0.24\textwidth}
    \centering
    \includegraphics[width=\linewidth]{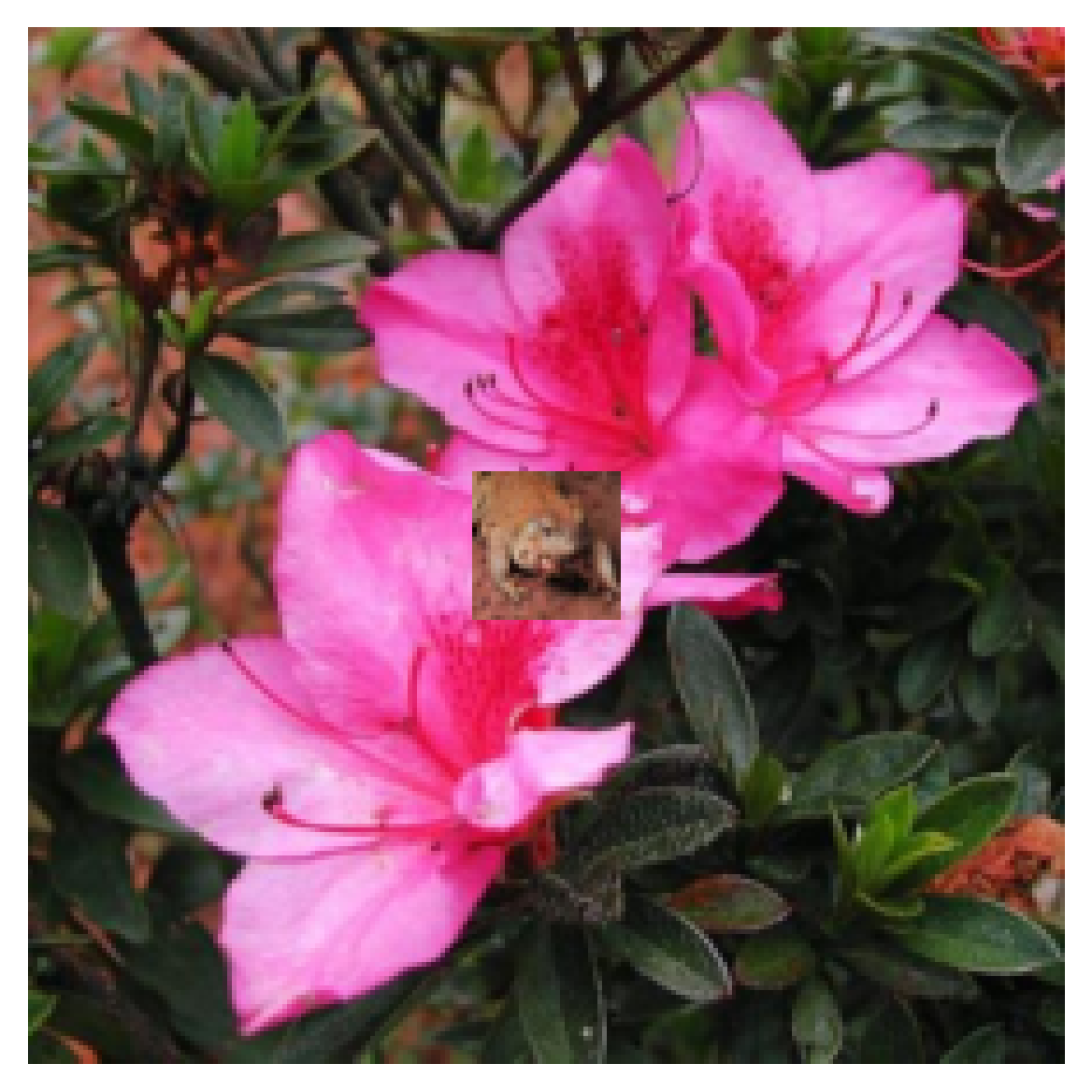}
    \caption{SBFP}
    \label{fig:sbfp}
\end{subfigure}
\hfill
\begin{subfigure}[b]{0.24\textwidth}
    \centering
    \includegraphics[width=\linewidth]{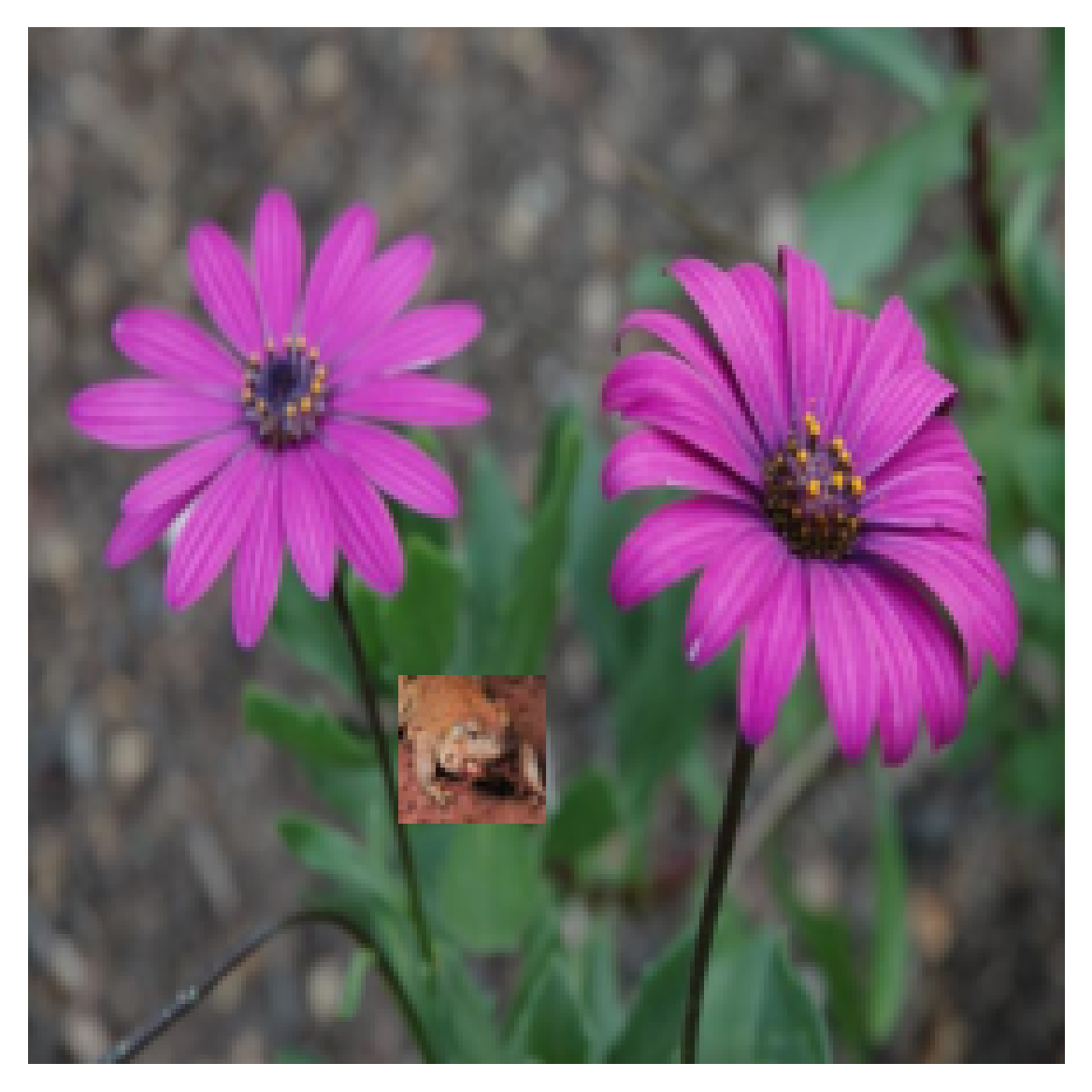}
    \caption{SBRP}
    \label{fig:sbrp}
\end{subfigure}
\caption{Vision benchmark conditions. A class-bearing CIFAR-10 foreground is
placed at a fixed (FP) or random (RP) position over a Gaussian random
background (RB) or a structured but task-irrelevant Flower102 background (SB).}
\label{fig:visdata}
\end{figure}

\begin{figure*}[!t]
\centering
\begin{subfigure}[b]{0.31\textwidth}
    \centering
    \includegraphics[width=\linewidth,trim=7mm 8mm 16mm 2.5mm,clip]{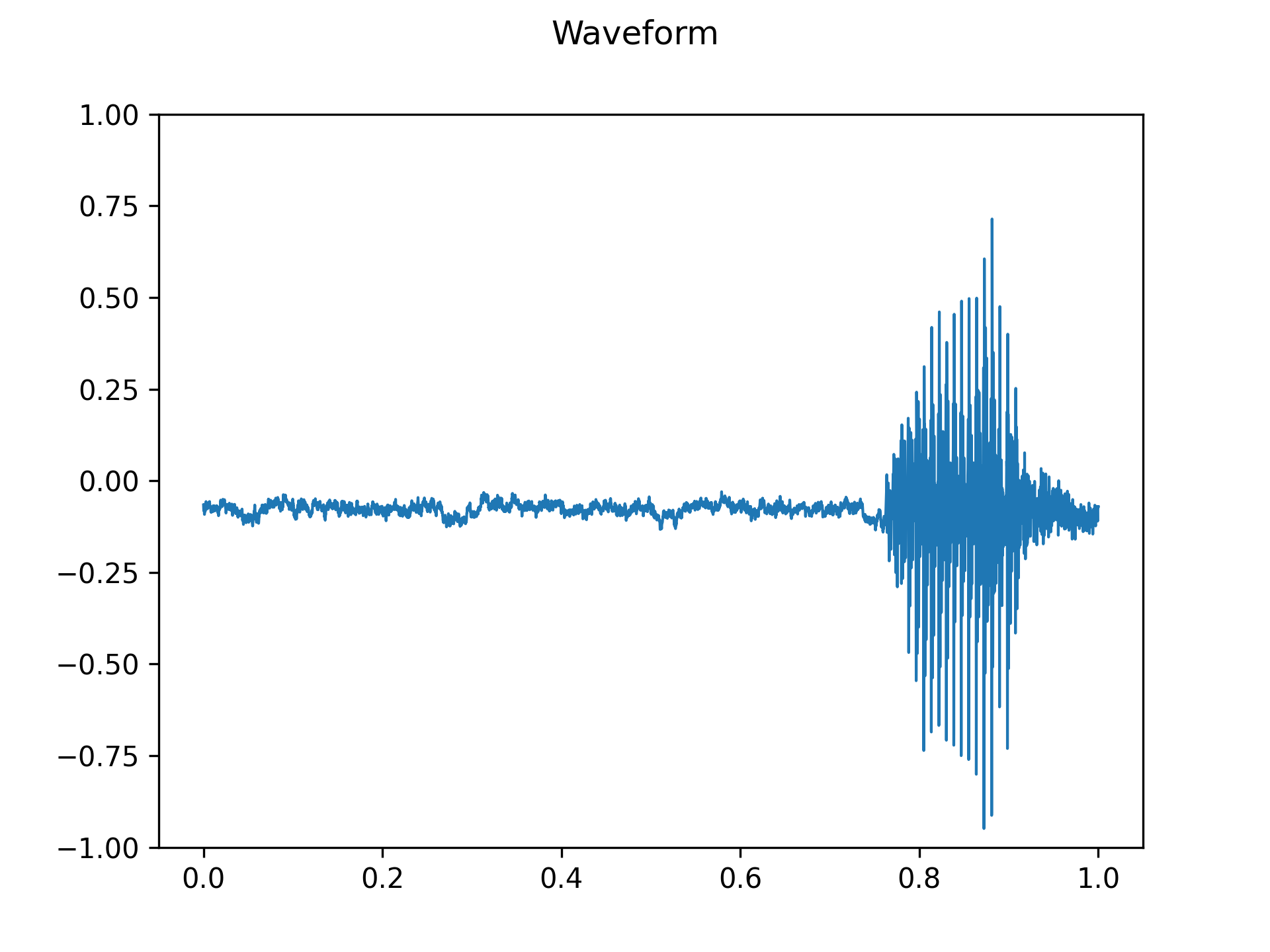}
    \caption{Command}
    \label{fig:command}
\end{subfigure}
\hfill
\begin{subfigure}[b]{0.31\textwidth}
    \centering
    \includegraphics[width=\linewidth,trim=7mm 8mm 16mm 2.5mm,clip]{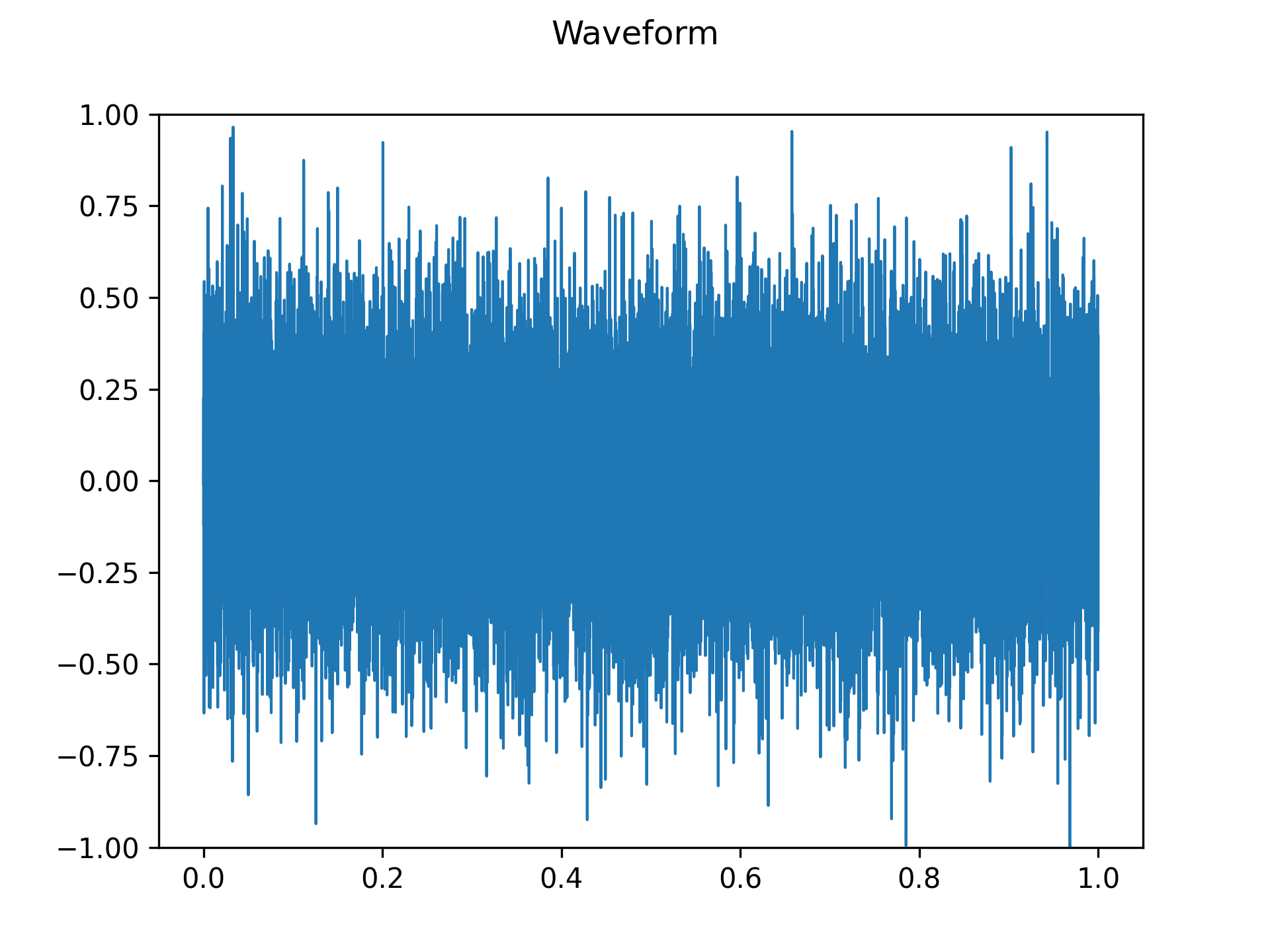}
    \caption{Gaussian}
    \label{fig:audio_noise}
\end{subfigure}
\hfill
\begin{subfigure}[b]{0.31\textwidth}
    \centering
    \includegraphics[width=\linewidth,trim=7mm 8mm 16mm 2.5mm,clip]{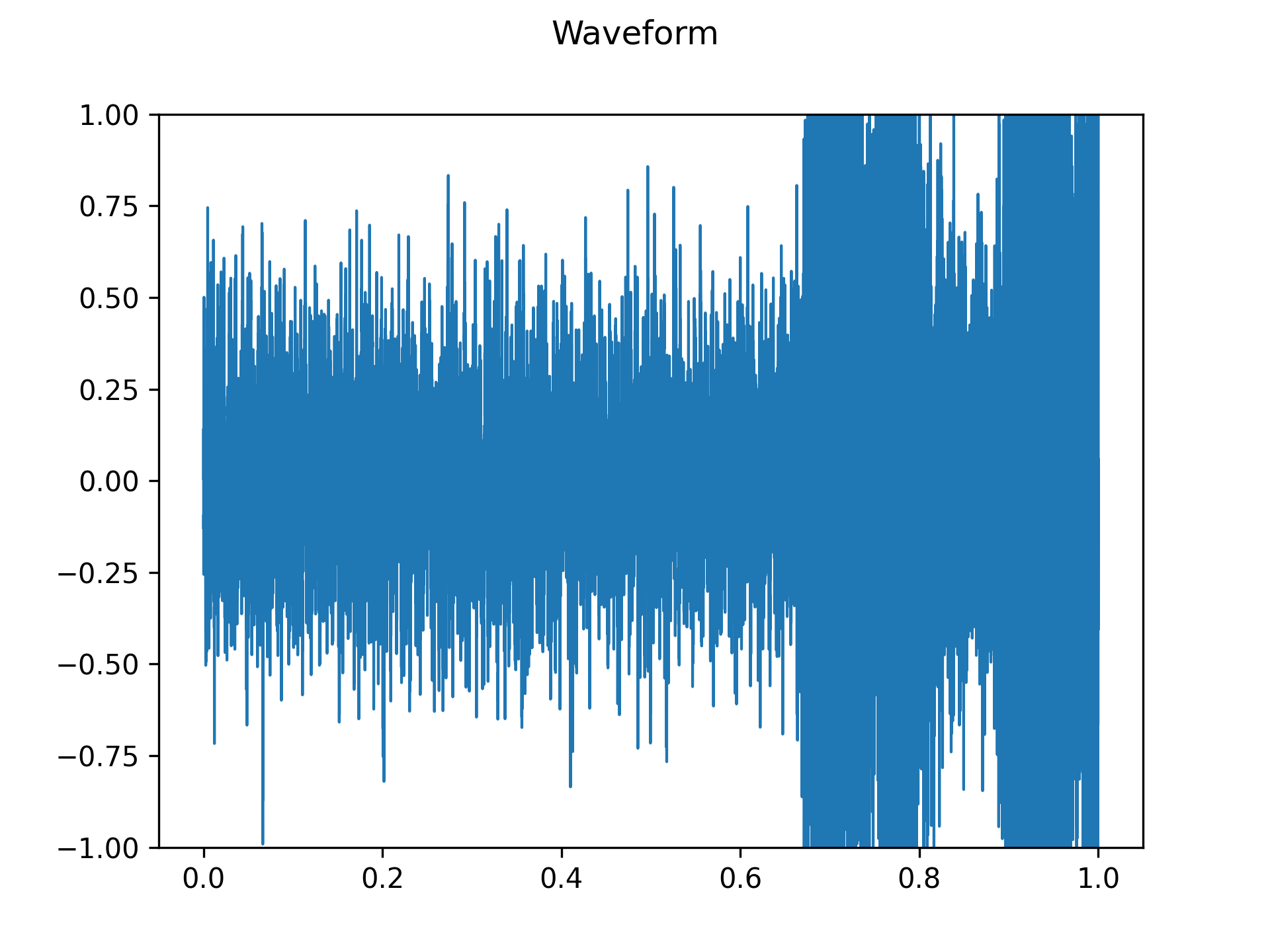}
    \caption{Environmental}
    \label{fig:forest}
\end{subfigure}
\caption{Audio sources: (a) the predictive speech command, (b) Gaussian input,
and (c) a structured but task-irrelevant environmental recording.}
\label{fig:auddata}
\end{figure*}

\paragraph{Symbolic tabular data.}\label{para:func}
The symbolic benchmark contains 15 human-designed nonlinear regression
functions spanning polynomial, trigonometric, exponential, and logarithmic
operations. Their informative-set sizes range from 1 to 10. Inputs and targets
are generated as

\begin{equation}
\begin{aligned}
x_i &\sim \mathcal{N}(0,1), \\
y &= f_j\!\left(x_{S^{\star}}\right)+\eta,
\qquad j\in\{1,\ldots,15\},
\end{aligned}
\label{eq:symbolic-generation}
\end{equation}

where variables appearing in \(f_j\) define \(S^{\star}\), \(\eta\) is used
only in controlled label-noise sweeps, and every additional coordinate is
task-irrelevant. The complete function catalog appears in
\appendixlocation{sec:appendix_symbolic_functions}. Experiments vary the
irrelevant-variable count, data size, label noise, model configuration, and
optimizer.

\paragraph{Vision data.}\label{para:vis}
Each \(224\times224\) input contains a \(32\times32\) CIFAR-10 foreground
\cite{krizhevsky2009learning}, whose class is the prediction target. The
background is Gaussian input or an unrelated Flower102 image
\cite{Nilsback08}. The four configurations are RBFP, RBRP, SBFP, and SBRP;
each stores the image identifier, foreground class, and foreground coordinates
in \texttt{meta\_data.csv}. Fixed-position conditions provide common spatial
support, whereas random-position conditions require instance-specific
localization and are reported separately because training is less reliable.

\paragraph{Audio data.}\label{para:aud}
Each sample is a ten-channel, one-second waveform at 16\,kHz. One channel
contains one of 35 Speech Commands classes \cite{warden2018speech}; the other
nine contain Gaussian inputs or unrelated Rainforest Connection Species audio
\cite{rfcx-species-audio-detection}. The RBFP, RBRP, SBFP, and SBRP
configurations record the command class and predictive-channel position in
\texttt{meta\_data.csv}. Models did not converge when the predictive channel
varied, so RBRP and SBRP are documented but excluded from the principal
comparison.

\paragraph{Splits and training protocol.}
Vision uses 50,000 training and 10,000 validation images constructed from
disjoint source-dataset splits. Models are initialized from PyTorch pretrained
weights and fine-tuned for 30 epochs with batch size 128, AdamW at learning rate
\(10^{-3}\), and cosine annealing with warm restarts. Vision attribution maps
are smoothed with a Gaussian filter (\(\sigma=6\)); feature ablation uses 49
non-overlapping \(32\times32\) patches. Each retained audio condition contains
84,843 training and 9,981 validation waveforms normalized to \([-1,1]\). Audio
models are trained for 30 epochs with batch size 128; the common backbone begins
with a 1D convolution and uses three hidden layers of width 60 with temporal max
pooling. All experiments use PyTorch and were run on an RTX 3080.

\begin{table*}[!t]
  \caption{Benchmark design, evaluation, and ground-truth annotation.}
  \label{tab:data}
  \centering
  \scriptsize
  \setlength{\tabcolsep}{3pt}
  \resizebox{\textwidth}{!}{%
  \begin{tabular}{llllll}
    \toprule
    Data & Task & Predictive performance & Informative-set identification & Ground-truth informative set & Task-irrelevant inputs \\
    \midrule
    Symbolic & Regression & NPS & RRA & Variables in the generating formula & Additional independent variables \\
    Vision & Classification & Accuracy & IoU & CIFAR-10 foreground region & Gaussian noise or Flower102 background \\
    Audio & Classification & Accuracy & RMA & Speech Commands channel & Gaussian noise or rainforest-audio channels \\
    \bottomrule
  \end{tabular}
  }
\end{table*}

\subsection{Dataset Release and Reproducibility}

The benchmark is distributed through \datasetrepo{Hugging Face} under the
persistent identifier \datasetdoi{doi:10.57967/hf/2482}, with curation and
evaluation code in \projectrepo{the project repository}. The release contains
the symbolic configuration and four vision and four audio configurations.
Symbolic features and targets use CSV; images use PNG; waveforms use WAV; and
CSV metadata provide labels and ground-truth informative locations or channels.
The symbolic data provide a training split, while vision and audio provide
training and validation splits. The release uses CC BY-NC 4.0, and maintenance
and troubleshooting are handled through the \datasetdiscussion{dataset
community page}.

\subsection{Evaluation Metrics}

We report three complementary evaluation dimensions separately: predictive
performance, informative-set identification, and operational properties. No
aggregate score is formed.

\paragraph{Normalized Prediction Similarity.}
For symbolic regression, we first define the mean absolute error (MAE) as

\begin{equation}\label{eq:mae}
    \operatorname{MAE}
    =
    \frac{1}{N}\sum_{i=1}^{N}
    |\hat{y}_i-y_i|.
\end{equation}

The symbolic tasks include polynomial, trigonometric, and exponential target
functions, for which raw MAE can vary substantially with target scale. We
therefore normalize the error using symmetric mean absolute percentage error
(sMAPE):

\begin{equation}\label{eq:smape}
    \operatorname{sMAPE}
    =
    \frac{1}{N}\sum_{i=1}^{N}
    \frac{2|\hat{y}_i-y_i|}
    {|\hat{y}_i|+|y_i|+\epsilon},
\end{equation}

where \(\epsilon=10^{-8}\) prevents an undefined zero denominator. To make
the normalized results directly interpretable in a high-is-better plot, we
report the \emph{Normalized Prediction Similarity} (NPS):

\begin{equation}\label{eq:6}
    \operatorname{NPS}
    = 1-\frac{1}{2}\operatorname{sMAPE}
    = \frac{1}{N}\sum_{i=1}^{N}
    \left(
    1-\frac{|\hat{y}_i-y_i|}
    {|\hat{y}_i|+|y_i|+\epsilon}
    \right).
\end{equation}

NPS is scale-free and lies in \([0,1]\), with one denoting an exact match.
It permits results from the heterogeneous symbolic tasks to be summarized in
one plot without raw-MAE scale dominating the comparison. Vision and audio use
classification accuracy.

\paragraph{Relevance Rank Accuracy.}
For symbolic data, let \(a\) be an evaluated attribution vector and let
\(\operatorname{Top}_{k}(|a|)\) contain its \(k=|S^{\star}|\) largest
attribution magnitudes. Relevance Rank Accuracy (RRA) is

\begin{equation}\label{eq:7}
    \operatorname{RRA}(a,S^{\star})
    =
    \frac{|\operatorname{Top}_{k}(|a|)\cap S^{\star}|}{k}.
\end{equation}

This is precision at \(k\), and because both sets have size \(k\), it is also
recall at \(k\). For a collection of local explanations, the reported RRA is
the mean of the instance-level top-\(k\) overlap scores across test samples.
Under a uniformly random ranking, expected RRA is \(k/d\); this baseline
therefore decreases as irrelevant variables are added.
RRA lies in \([0,1]\), with larger
values indicating better informative-set identification. Thus, for the
symbolic sweeps, NPS and RRA share the same bounded, larger-is-better scale,
which makes their relationship directly comparable in Figure~\ref{fig:simexp}.
RRA measures rank-based set overlap and does not evaluate ordering outside the selected set
\cite{arras2022clevrxai}.

\paragraph{Vision intersection over union.}
Vision attributions are converted to recovered spatial regions at the same
evaluation granularity as the foreground annotation. We select the \(k\) pixels
with largest absolute attribution, where \(k\) equals the foreground area, and
take their minimum enclosing axis-aligned rectangle. We report intersection
over union (IoU) between this recovered rectangle and the annotated foreground
rectangle. IoU is retained as a modality-specific localization measure rather
than forced into the rank-based tabular metric.

\paragraph{Audio Relevance Mass Accuracy.}
For an audio attribution tensor \(a^{(i)}\in\mathbb{R}^{C\times T}\) and the
informative channel \(\hat c\), the existing audio experiments measure the
fraction of absolute attribution magnitude assigned to that channel:

\begin{equation}\label{eq:rma}
    \operatorname{RMA}
    =\frac{1}{N}\sum_{i=1}^{N}
    \frac{\sum_{t=1}^{T}|a^{(i)}_{\hat c,t}|}
    {\sum_{c=1}^{C}\sum_{t=1}^{T}|a^{(i)}_{c,t}|+\epsilon}.
\end{equation}

This quantity is Relevance Mass Accuracy (RMA) applied to a channel mask
\cite{arras2022clevrxai}. This interpretation retains the reported numerical
values without recomputation.

\paragraph{Stability and computational cost.}
The symbolic experiments additionally report agreement of top-ranked features
across samples, convergence during model training, runtime, and memory. These
are complementary evaluation dimensions rather than components of a composite
metric or a formal trade-off score.

\section{Experiments}\label{sec:exp}

We organize the experiments around three questions: how irrelevant inputs and
predictive performance affect attribution; whether attribution complexity
improves identification, stability, or efficiency; and whether tabular
observations extend to spatial and channel-structured inputs.

\subsection{Symbolic Data: Irrelevant Features and Model Performance}
\definecolor{DLcolor}{HTML}{22787D}
\definecolor{FAcolor}{HTML}{1C7A97}
\definecolor{IGcolor}{HTML}{4579A9}
\definecolor{KScolor}{HTML}{7773AD}
\definecolor{LIMEcolor}{HTML}{A26AA0}
\definecolor{SAcolor}{HTML}{BF6484}
\begin{figure*}[!t]
\centering

\begin{subfigure}[b]{0.24\textwidth}
    \centering
    \includegraphics[width=\linewidth,height=1.65in]{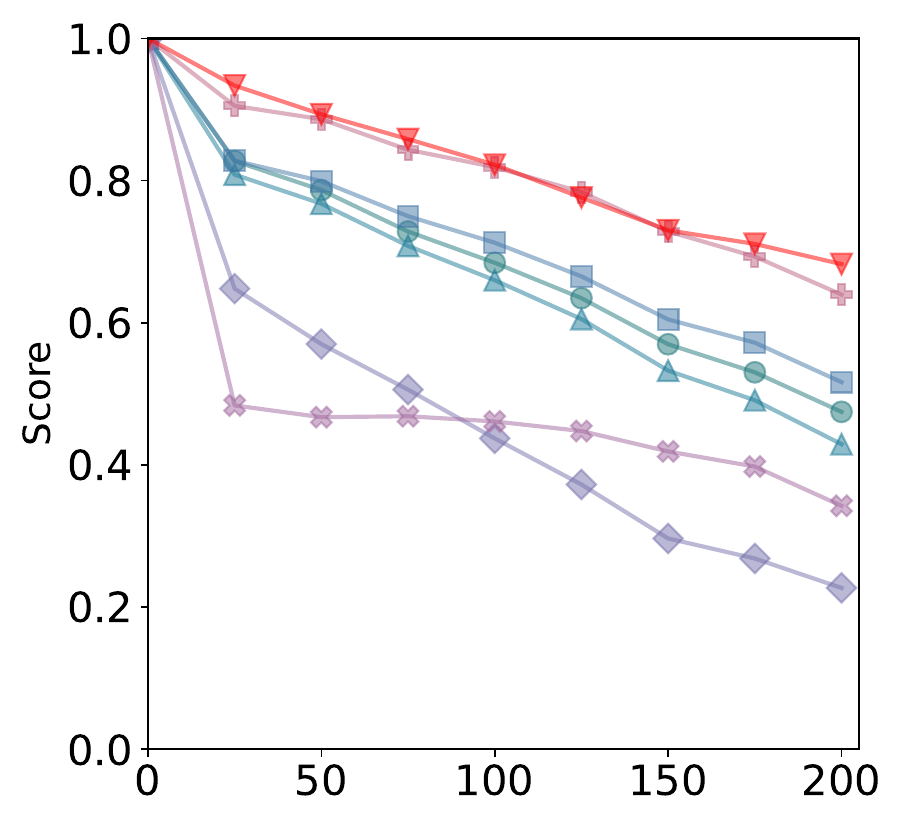}
    \caption{Noisy features}
    \label{fig:nnf}
\end{subfigure}\hfill
\begin{subfigure}[b]{0.24\textwidth}
    \centering
    \includegraphics[width=\linewidth,height=1.65in]{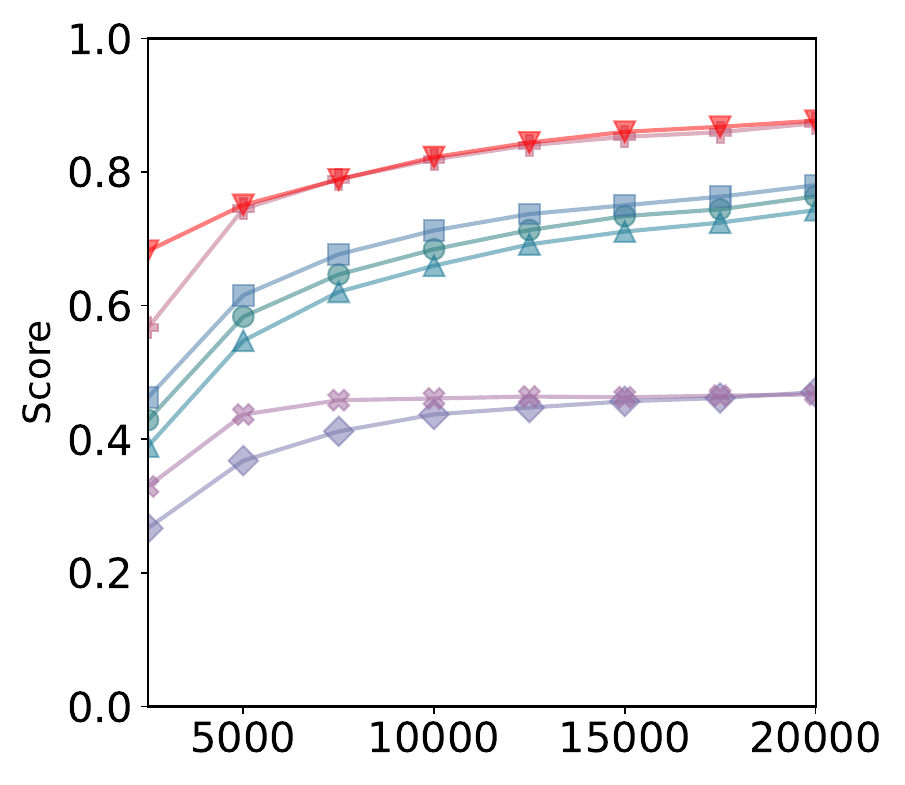}
    \caption{Training data}
    \label{fig:nd}
\end{subfigure}\hfill
\begin{subfigure}[b]{0.24\textwidth}
    \centering
    \includegraphics[width=\linewidth,height=1.65in]{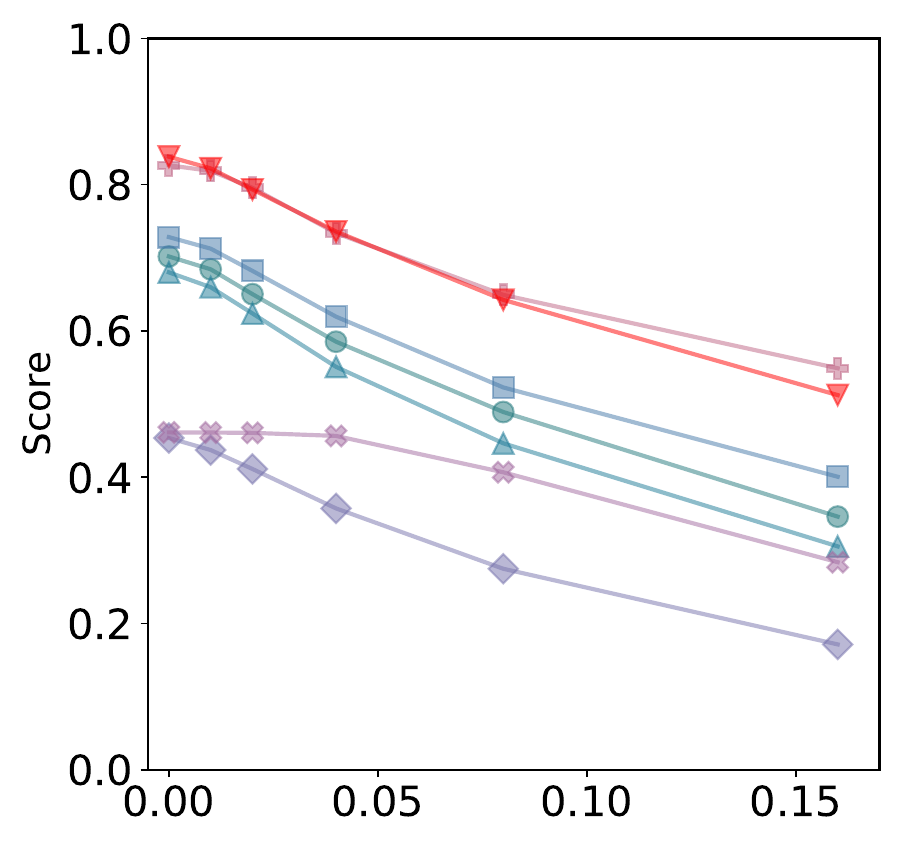}
    \caption{Label noise}
    \label{fig:ln}
\end{subfigure}\hfill
\begin{subfigure}[b]{0.24\textwidth}
    \centering
    \includegraphics[width=\linewidth,height=1.65in]{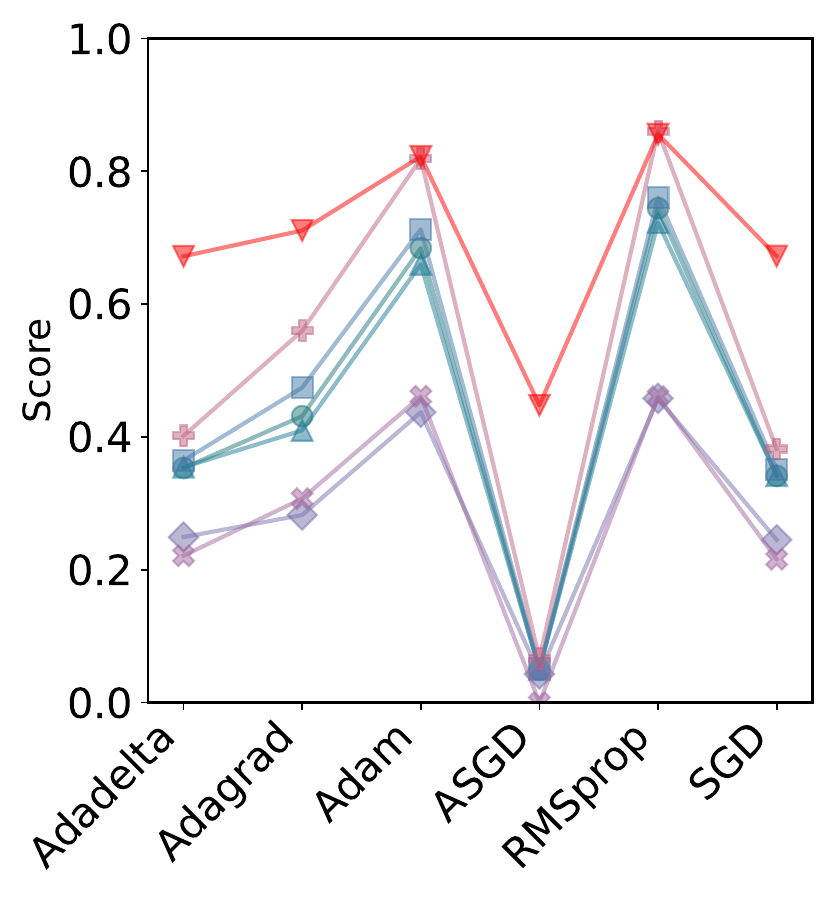}
    \caption{Optimizers}
    \label{fig:optimizer}
\end{subfigure}

\vspace{-2pt}

\begin{subfigure}[b]{0.24\textwidth}
    \centering
    \includegraphics[width=\linewidth,height=1.65in]{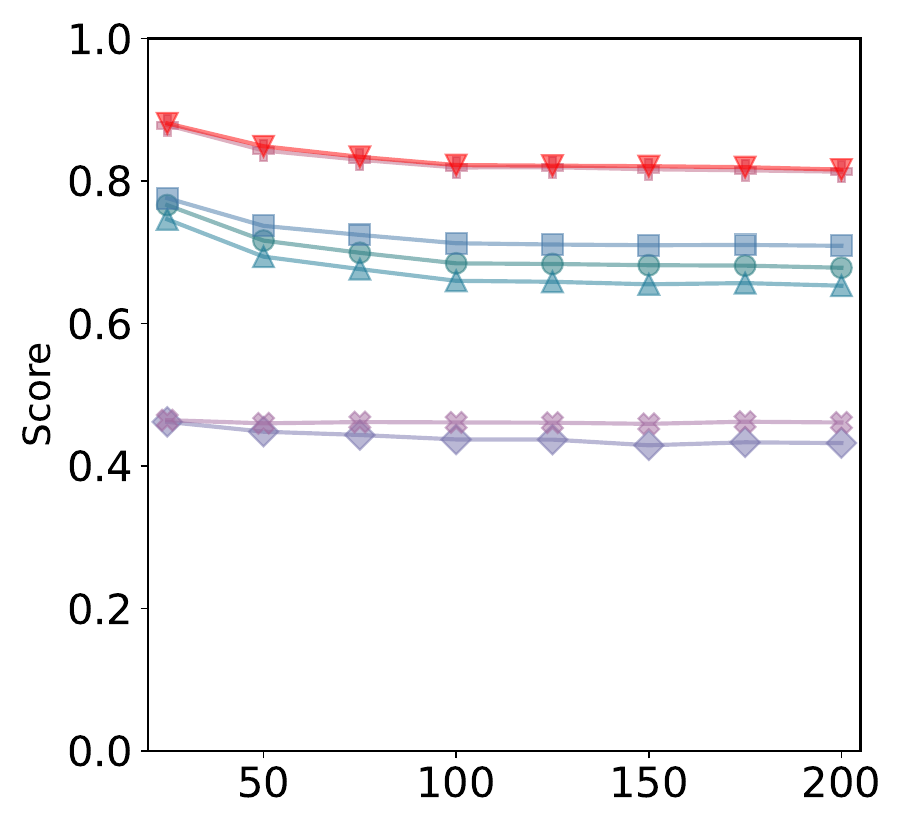}
    \caption{Model width}
    \label{fig:widths}
\end{subfigure}\hfill
\begin{subfigure}[b]{0.24\textwidth}
    \centering
    \includegraphics[width=\linewidth,height=1.65in]{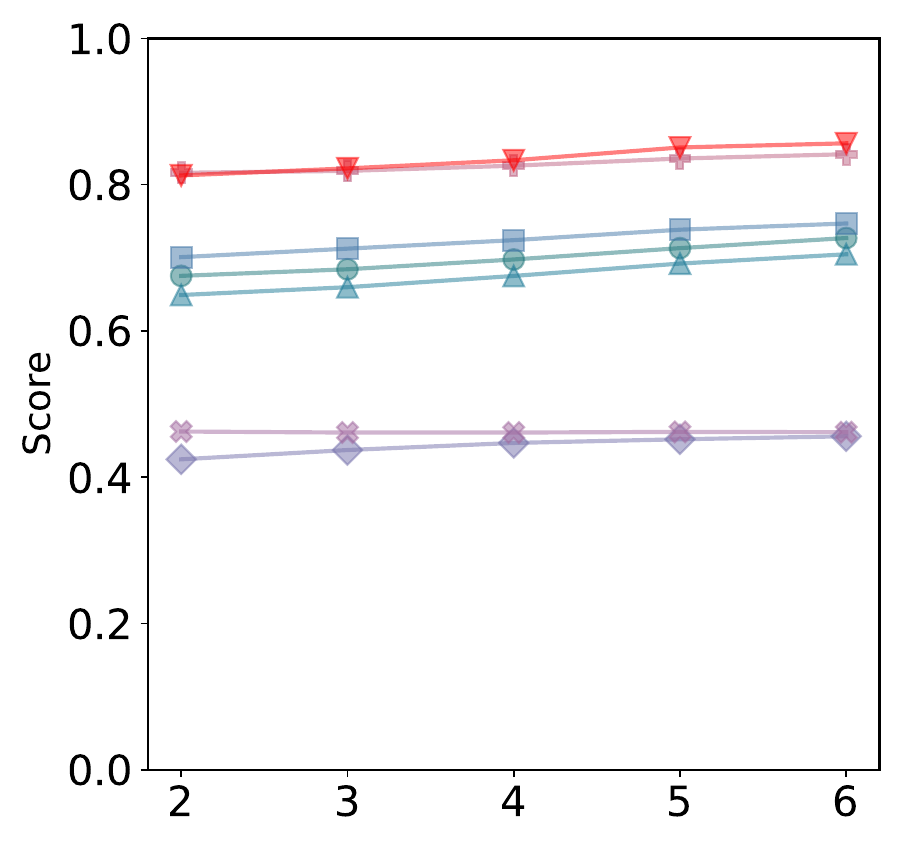}
    \caption{Model depth}
    \label{fig:depths}
\end{subfigure}\hfill
\begin{subfigure}[b]{0.24\textwidth}
    \centering
    \includegraphics[width=\linewidth,height=1.65in]{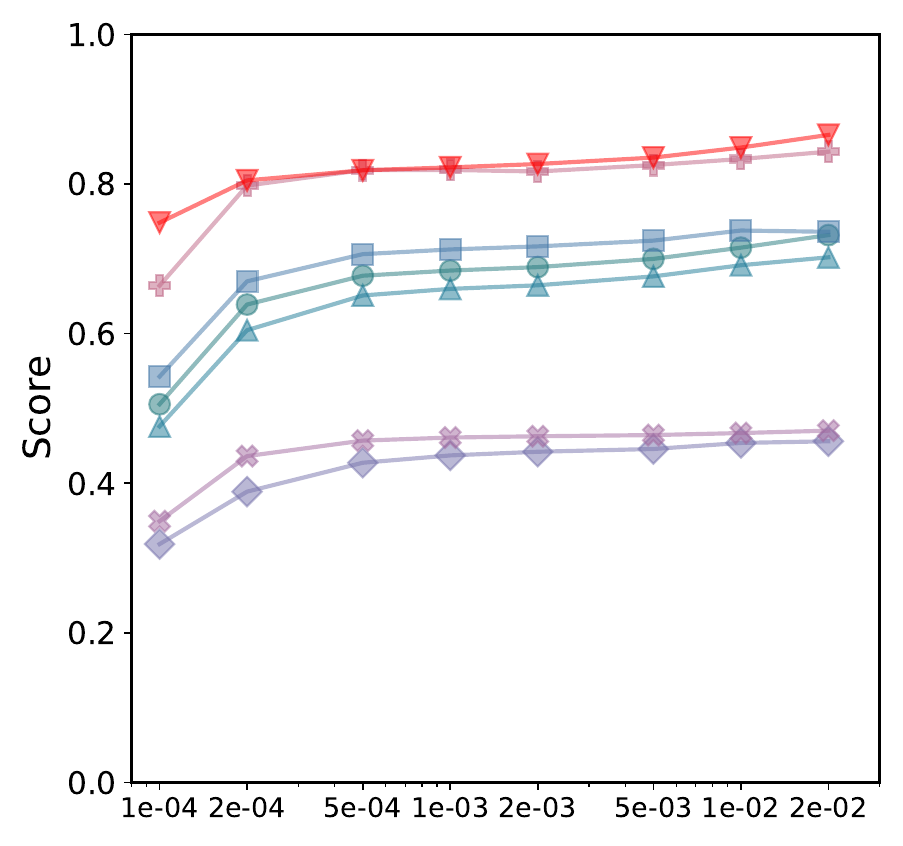}
    \caption{Learning rate}
    \label{fig:lr}
\end{subfigure}\hfill
\begin{subfigure}[b]{0.24\textwidth}
    \centering
    \includegraphics[width=\linewidth,height=1.65in]{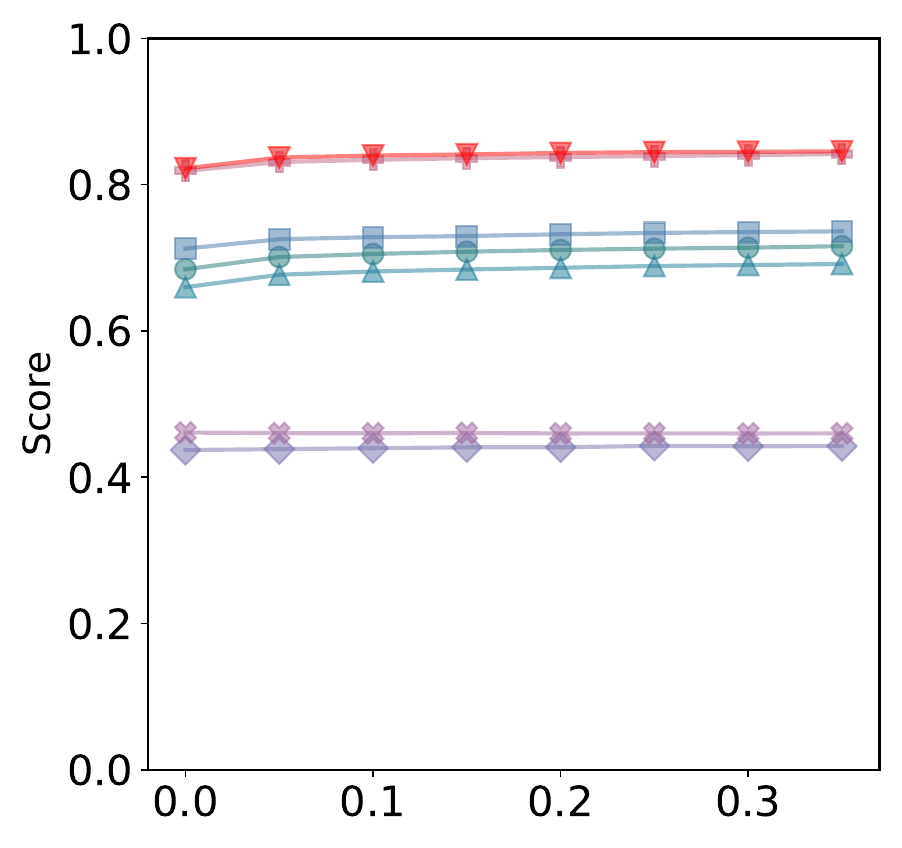}
    \caption{Dropout rate}
    \label{fig:dr}
\end{subfigure}

\caption{Symbolic one-factor sweeps. Colors and markers identify the six methods:
\textcolor{DLcolor}{$\bullet$} DL,
\textcolor{FAcolor}{$\blacktriangle$} FA,
\textcolor{IGcolor}{$\blacksquare$} IG,
\textcolor{KScolor}{$\blacklozenge$} KS,
\textcolor{LIMEcolor}{$\boldsymbol{\times}$} LIME, and
\textcolor{SAcolor}{$\oplus$} SA. The \textcolor{red}{$\blacktriangledown$}
denotes prediction NPS. NPS and RRA are normalized to \([0,1]\), with larger
values indicating better performance.}
\label{fig:simexp}
\end{figure*}

\begin{figure*}[!t]
\centering

\begin{subfigure}[b]{0.24\textwidth}
    \centering
    \includegraphics[width=\linewidth,height=1.60in]{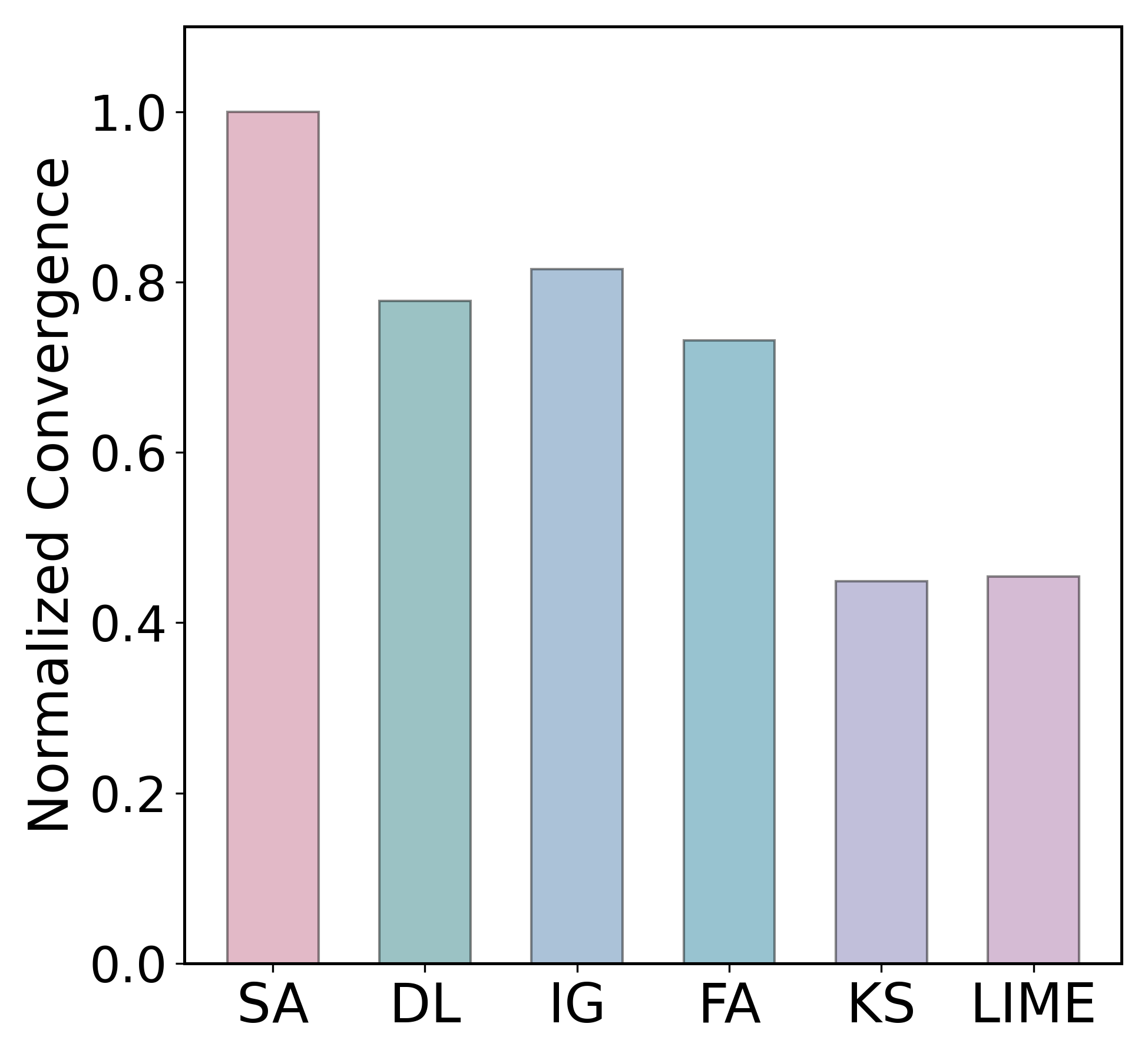}
    \caption{Convergence}
    \label{fig:convergence}
\end{subfigure}
\hfill
\begin{subfigure}[b]{0.24\textwidth}
    \centering
    \includegraphics[width=\linewidth,height=1.60in]{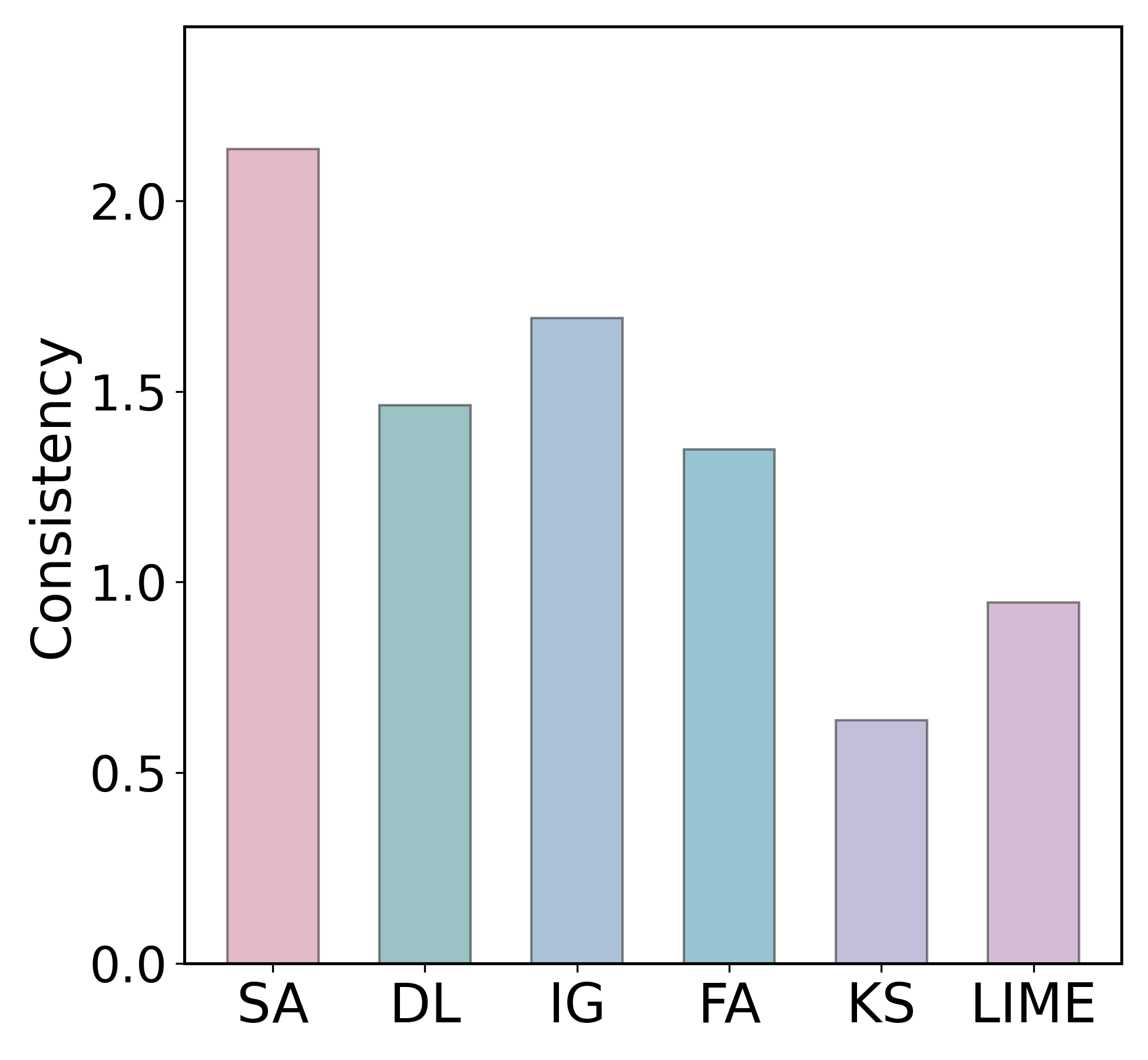}
    \caption{Consistency}
    \label{fig:consistency}
\end{subfigure}
\hfill
\begin{subfigure}[b]{0.24\textwidth}
    \centering
    \includegraphics[width=\linewidth,height=1.60in]{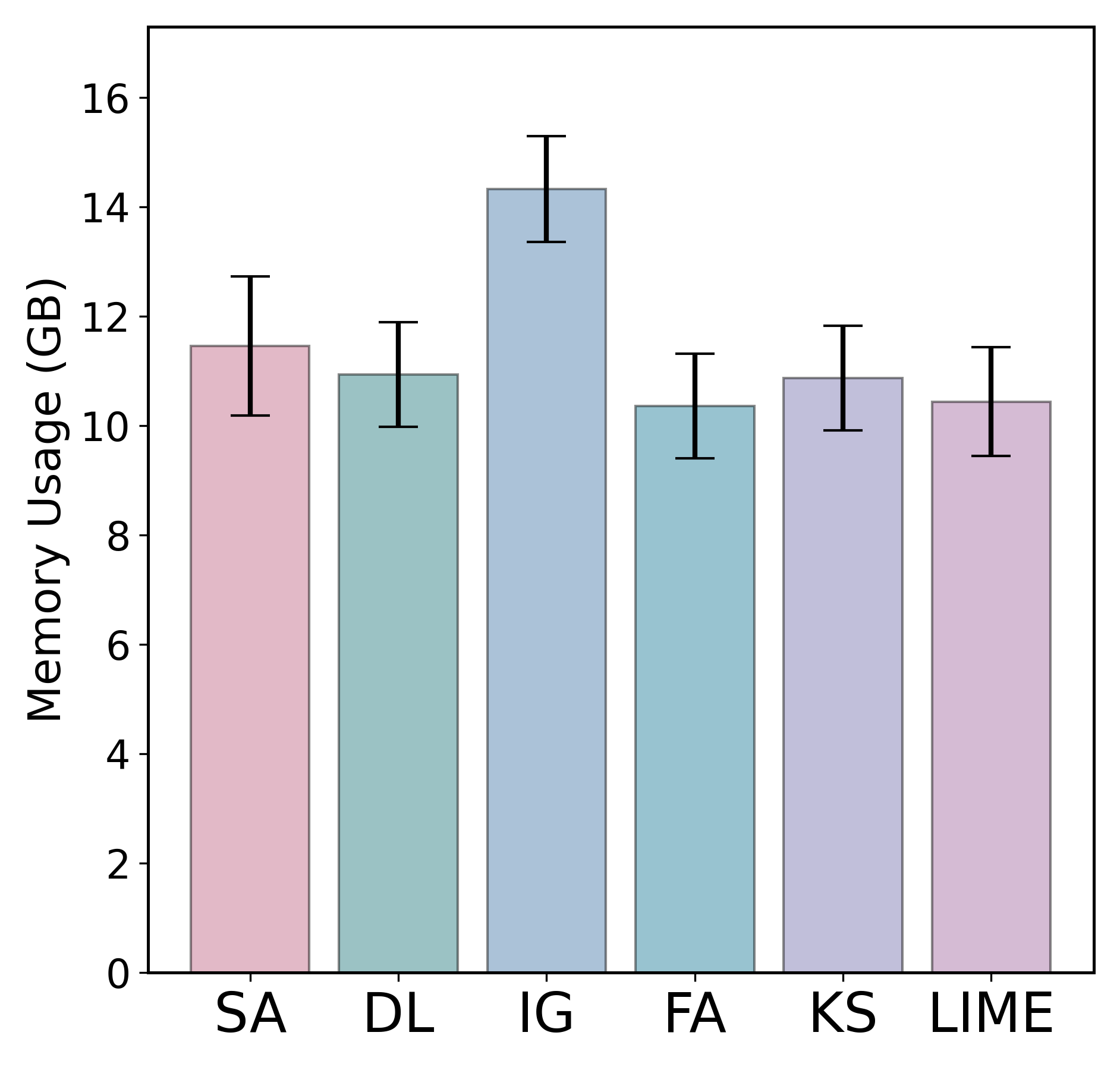}
    \caption{Memory cost}
    \label{fig:memory}
\end{subfigure}
\hfill
\begin{subfigure}[b]{0.24\textwidth}
    \centering
    \includegraphics[width=\linewidth,height=1.60in]{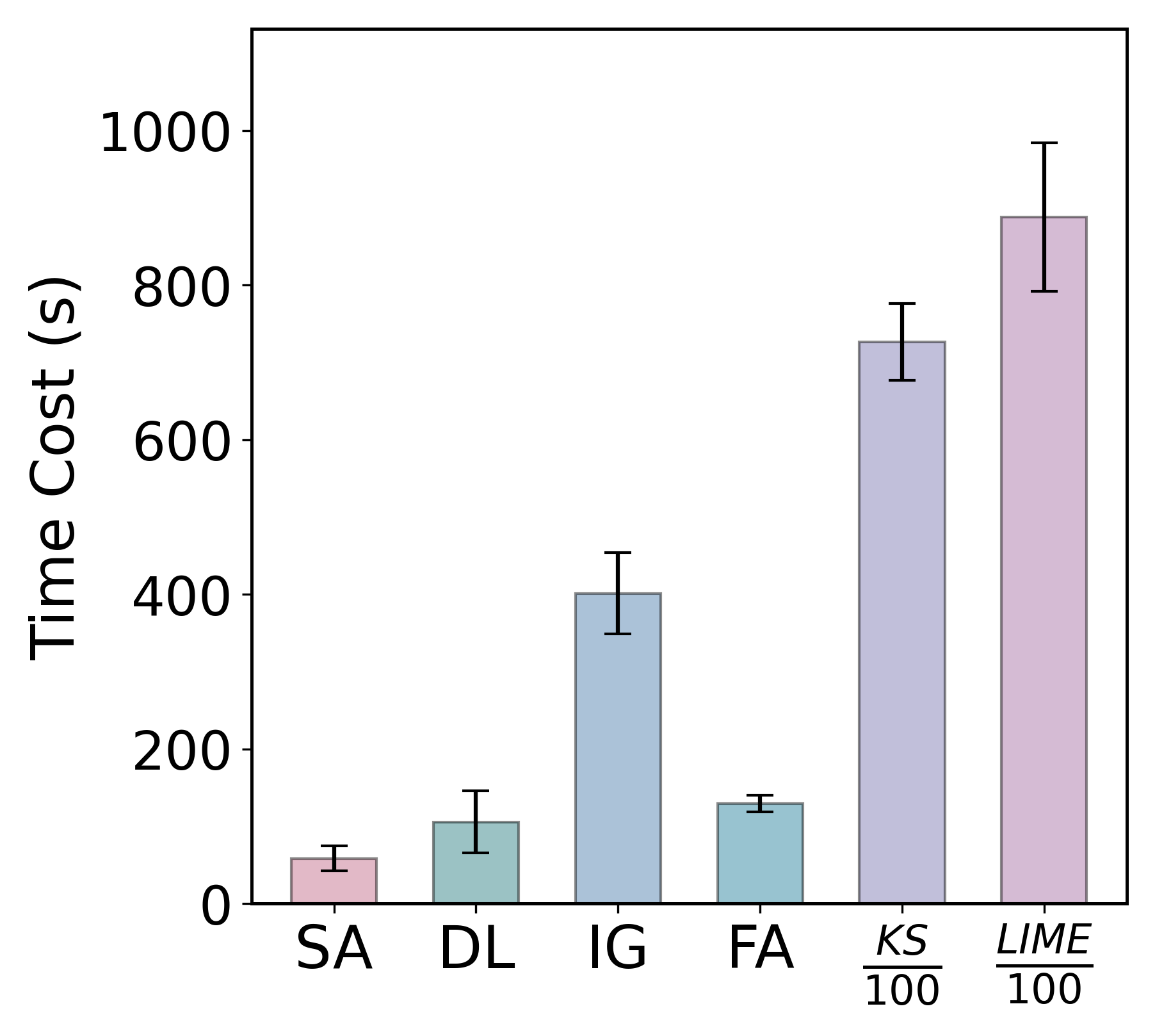}
    \caption{Time cost}
    \label{fig:time}
\end{subfigure}

\caption{Complementary symbolic-data evaluation. Convergence is the area under
the RRA curve over 300 training epochs; consistency compares each sample's
top-ranked features with the dataset-average attribution. Both are normalized
to SA. Memory and runtime are measured at test time. KS and LIME runtime values
are scaled by (1/100) for visualization.}
\label{fig:simulation}
\end{figure*}

Each sweep changes one component of a default configuration containing 10,000
samples, a 4:1 train--test split, 100 irrelevant variables, label-noise scale
0.01, an MLP of width 100 and depth 3, Adam with learning rate 0.001, no
dropout, and 1,000 training epochs. Figure~\ref{fig:simexp} reports the
model's held-out NPS together with each method's RRA.

Across most one-factor sweeps, predictive performance and attribution
identification change in the same direction, although exceptions occur for
particular methods and hyperparameters. Increasing the number of irrelevant
variables or label noise generally reduces both NPS and RRA, whereas additional
training data or training epochs generally improves both. Model and
optimization choices also matter, but their effects are not monotone: greater
width does not necessarily improve held-out prediction, and an optimizer that
fails to fit the task cannot support meaningful attribution.
Because NPS and RRA both lie in \([0,1]\) and are larger-is-better, their
co-variation is directly comparable in Figure~\ref{fig:simexp}. These results
motivate reporting prediction and attribution metrics together rather than
interpreting attribution scores in isolation.

No method dominates every sweep. Saliency is frequently among the strongest
rank-based results, but exceptions occur across configurations and methods.
Within each sweep, however, the relative ordering of the methods is largely
stable as NPS changes, even as their absolute RRA values rise or fall with
predictive performance. Figure~\ref{fig:simulation} provides a multi-criteria
view rather than a single ranking. Saliency has high cross-sample
consistency and the lowest measured overhead, whereas path- or
perturbation-based procedures can require substantially more time or memory.
These measurements support a practical efficiency advantage for simple
gradients, not a universal claim of superior attribution quality.

\subsection{Vision: Fixed Spatial Support and Position Sensitivity}

\begin{table*}[!t]
  \centering
  \caption{Vision accuracy and IoU under fixed-position random-background
  (RBFP) and structured-background (SBFP) conditions.}
  \label{tab:vision_fixed}
  \scriptsize
  \renewcommand{\arraystretch}{0.76}
  \setlength{\tabcolsep}{2.1pt}
  \resizebox{0.8\textwidth}{!}{%
  \begin{tabular}{lccccccccc}
    \toprule
    \multicolumn{10}{l}{\textbf{Panel A: Random background, fixed position (RBFP)}} \\
    Architecture & Acc. (\%) & SA & DL & IG & FA & KS & DLS & GB & LIME \\
    \midrule
    ConvNeXt-Tiny~\cite{liu2022convnet} & \(85.11\) & \(0.829\) & \(0.666\) & \(0.554\) & \(\mathbf{0.860}\) & \(0.021\) & \(0.666\) & \(0.656\) & \(0.021\) \\
    DenseNet-121~\cite{huang2017densely} & \(83.38\) & \(0.738\) & \(0.566\) & \(0.601\) & \(\mathbf{0.748}\) & \(0.022\) & \(0.674\) & \(0.639\) & \(0.022\) \\
    EfficientNet-V2~\cite{tan2021efficientnetv2} & \(85.07\) & \(0.791\) & \(0.611\) & \(0.636\) & \(\mathbf{0.866}\) & \(0.021\) & \(0.611\) & \(0.549\) & \(0.021\) \\
    ResNet-18~\cite{he2016deep} & \(77.44\) & \(\mathbf{0.678}\) & \(0.530\) & \(0.551\) & \(0.677\) & \(0.020\) & \(0.532\) & \(0.450\) & \(0.020\) \\
    ResNet-50~\cite{he2016deep} & \(83.23\) & \(0.656\) & \(0.326\) & \(0.457\) & \(\mathbf{0.682}\) & \(0.022\) & \(0.513\) & \(0.585\) & \(0.022\) \\
    Swin-T~\cite{liu2021swin} & \(78.10\) & \(\mathbf{0.827}\) & \(0.657\) & \(0.625\) & \(0.547\) & \(0.021\) & \(0.658\) & \(0.639\) & \(0.021\) \\
    VGG-16~\cite{simonyan2014very} & \(10.00\) & \(0.021\) & \(0.000\) & \(0.000\) & \(0.021\) & \(0.021\) & \(0.000\) & \(0.021\) & \(0.021\) \\
    ViT-B/16~\cite{dosovitskiy2020image} & \(48.05\) & \(0.285\) & \(0.148\) & \(0.117\) & \(\mathbf{0.369}\) & \(0.115\) & \(0.111\) & \(0.047\) & \(0.115\) \\
    ViT-L/32~\cite{dosovitskiy2020image} & \(53.27\) & \(0.701\) & \(0.273\) & \(0.269\) & \(\mathbf{0.709}\) & \(0.101\) & \(0.257\) & \(0.193\) & \(0.101\) \\
    \bottomrule
  \end{tabular}
  }
  \vspace{1pt}
  \resizebox{0.8\textwidth}{!}{%
  \begin{tabular}{lccccccccc}
    \toprule
    \multicolumn{10}{l}{\textbf{Panel B: Structured background, fixed position (SBFP)}} \\
    Architecture & Acc. (\%) & SA & DL & IG & FA & KS & DLS & GB & LIME \\
    \midrule
    ConvNeXt-Tiny~\cite{liu2022convnet} & \(10.00\) & \(0.021\) & \(0.028\) & \(0.029\) & \(0.069\) & \(\mathbf{0.115}\) & \(0.032\) & \(0.002\) & \(\mathbf{0.115}\) \\
    DenseNet-121~\cite{huang2017densely} & \(85.60\) & \(0.767\) & \(0.645\) & \(0.658\) & \(\mathbf{0.857}\) & \(0.022\) & \(0.668\) & \(0.036\) & \(0.022\) \\
    EfficientNet-V2~\cite{tan2021efficientnetv2} & \(85.91\) & \(0.743\) & \(0.535\) & \(0.640\) & \(\mathbf{0.898}\) & \(0.022\) & \(0.556\) & \(0.036\) & \(0.022\) \\
    ResNet-18~\cite{he2016deep} & \(80.41\) & \(0.745\) & \(0.610\) & \(0.618\) & \(\mathbf{0.825}\) & \(0.021\) & \(0.631\) & \(0.037\) & \(0.021\) \\
    ResNet-50~\cite{he2016deep} & \(87.71\) & \(0.762\) & \(0.643\) & \(0.680\) & \(\mathbf{0.863}\) & \(0.023\) & \(0.676\) & \(0.651\) & \(0.023\) \\
    Swin-T~\cite{liu2021swin} & \(10.00\) & \(0.021\) & \(0.044\) & \(0.052\) & \(0.025\) & \(0.077\) & \(0.065\) & \(0.075\) & \(0.077\) \\
    VGG-16~\cite{simonyan2014very} & \(10.00\) & \(0.021\) & \(0.021\) & \(0.021\) & \(0.021\) & \(0.093\) & \(0.021\) & \(0.004\) & \(0.193\) \\
    ViT-B/16~\cite{dosovitskiy2020image} & \(52.50\) & \(\mathbf{0.671}\) & \(0.364\) & \(0.485\) & \(0.552\) & \(0.023\) & \(0.620\) & \(0.578\) & \(0.023\) \\
    ViT-L/32~\cite{dosovitskiy2020image} & \(53.78\) & \(\mathbf{0.723}\) & \(0.273\) & \(0.375\) & \(0.542\) & \(0.119\) & \(0.335\) & \(0.608\) & \(0.119\) \\
    \bottomrule
  \end{tabular}
  }
  \vspace{1pt}
  \begin{minipage}{0.8\textwidth}
    \scriptsize \textit{Note:} Entries are mean IoU; bold marks the highest
    reported attribution score for each architecture.
  \end{minipage}
\end{table*}

Table~\ref{tab:vision_fixed} reports the principal fixed-position comparison,
in which the informative region occupies common coordinates across samples.
Among non-collapsed classifiers, Saliency (SA) is consistently competitive: it
is first or second for DenseNet-121 and ResNet-18 in both backgrounds, and is
also the leading score for ViT-L/32 under SBFP. Feature Ablation is likewise
strong in several rows, but no method dominates every architecture or
background.

Near-chance classifier accuracy requires a different interpretation. For a
\(C\)-class task, accuracy near \(1/C\) indicates predictions that are
effectively uniform over classes and, in this controlled setting, a failed
optimization run rather than a learned classifier. For example, the 10\% rows
for VGG-16 under RBFP and for ConvNeXt-Tiny, Swin-T, and VGG-16 under SBFP are
at the CIFAR-10 chance level. Their attribution scores are therefore not
evidence of informative-region recovery. This distinction is especially clear
for VGG-16: SA is \(0.021\), close to the uniform allocation over 49 image
patches (\(1/49\)), whereas KS or LIME can report substantially larger scores.
Such departures from the uniform attribution baseline reflect method-specific
inductive biases when the predictor itself has collapsed, not reliable
localization.

RBRP and SBRP are retained in the released benchmark but omitted from the
principal comparison: moving the foreground removes shared spatial
correspondence and requires instance-specific localization. Training often
collapses or achieves weak accuracy in these conditions, and the resulting
attribution maps are mostly uninformative or method-dependent.

\subsection{Audio: Fixed Informative Channel}

\begin{audiotable}
  \centering
  \caption{Audio accuracy and RMA under fixed-channel Gaussian (RBFP) and
  structured task-irrelevant (SBFP) backgrounds.}
  \label{tab:audio_fixed}
  \scriptsize
  \renewcommand{\arraystretch}{0.76}
  \setlength{\tabcolsep}{2.8pt}
  \resizebox{0.8\textwidth}{!}{%
  \begin{tabular}{lccccccccc}
    \toprule
    \multicolumn{10}{l}{\textbf{Panel A: Random background, fixed channel (RBFP)}} \\
    Architecture & Acc. (\%) & SA & DL & IG & FA & KS & DLS & GB & LIME \\
    \midrule
    Conformer~\cite{gulati2020conformer} & \(20.97\) & \(0.310\) & \(0.015\) & \(0.025\) & \(0.225\) & \(0.277\) & \(0.025\) & \(0.100\) & \(\mathbf{0.320}\) \\
    LSTM~\cite{sak2014long} & \(73.77\) & \(0.684\) & \(0.590\) & \(0.260\) & \(0.515\) & \(0.431\) & \(0.248\) & \(\mathbf{0.697}\) & \(0.564\) \\
    RNN~\cite{medsker2001recurrent} & \(67.00\) & \(0.653\) & \(0.585\) & \(0.275\) & \(0.447\) & \(0.427\) & \(0.440\) & \(\mathbf{0.814}\) & \(0.548\) \\
    TCN~\cite{kalchbrenner2016neural} & \(80.96\) & \(0.641\) & \(0.297\) & \(0.258\) & \(0.539\) & \(0.139\) & \(0.185\) & \(\mathbf{0.663}\) & \(0.325\) \\
    Transformer~\cite{vaswani2017attention} & \(23.25\) & \(\mathbf{0.460}\) & \(0.236\) & \(0.138\) & \(0.335\) & \(0.170\) & \(0.110\) & \(0.440\) & \(0.305\) \\
    Wav2Vec~2.0~\cite{baevski2020wav2vec} & \(3.68\) & \(0.100\) & \(0.024\) & \(0.039\) & \(0.101\) & \(0.101\) & \(0.042\) & \(0.206\) & \(0.101\) \\
    \midrule
    \multicolumn{10}{l}{\textbf{Panel B: Structured background, fixed channel (SBFP)}} \\
    Architecture & Acc. (\%) & SA & DL & IG & FA & KS & DLS & GB & LIME \\
    \midrule
    Conformer~\cite{gulati2020conformer} & \(3.85\) & \(0.100\) & \(0.014\) & \(0.019\) & \(0.100\) & \(0.019\) & \(0.019\) & \(0.100\) & \(0.099\) \\
    LSTM~\cite{sak2014long} & \(76.07\) & \(\mathbf{0.810}\) & \(0.772\) & \(0.439\) & \(0.546\) & \(0.378\) & \(0.291\) & \(0.770\) & \(0.624\) \\
    RNN~\cite{medsker2001recurrent} & \(63.16\) & \(0.644\) & \(0.541\) & \(0.340\) & \(0.482\) & \(0.517\) & \(0.410\) & \(\mathbf{0.821}\) & \(0.522\) \\
    TCN~\cite{kalchbrenner2016neural} & \(82.52\) & \(0.721\) & \(0.427\) & \(0.337\) & \(0.561\) & \(0.359\) & \(0.250\) & \(\mathbf{0.742}\) & \(0.350\) \\
    Transformer~\cite{vaswani2017attention} & \(25.75\) & \(\mathbf{0.513}\) & \(0.315\) & \(0.193\) & \(0.373\) & \(0.153\) & \(0.130\) & \(0.438\) & \(0.283\) \\
    Wav2Vec~2.0~\cite{baevski2020wav2vec} & \(3.68\) & \(0.100\) & \(0.021\) & \(0.038\) & \(0.100\) & \(0.038\) & \(0.039\) & \(0.198\) & \(0.100\) \\
    \bottomrule
  \end{tabular}
  }
  \vspace{1pt}
  \begin{minipage}{0.8\textwidth}
    \scriptsize \textit{Note:} Entries are mean RMA; bold marks the highest
    reported attribution score for each architecture.
  \end{minipage}
\end{audiotable}

Table~\ref{tab:audio_fixed} reports RMA for both fixed-channel backgrounds.
TCN attains the highest classification accuracy in both conditions. Across the
non-collapsed audio models, SA remains highly competitive: it is the highest
RMA for the Transformer in both panels and for the SBFP LSTM, and it is second
to GB for the RBFP LSTM, RNN, and TCN. The effect of replacing Gaussian inputs
with structured task-irrelevant audio is architecture dependent, so no
attribution method dominates the full comparison.

The low-accuracy rows again illustrate why attribution must be conditioned on
predictive performance. Accuracy near \(1/C\) is the random-classification
baseline for a \(C\)-class problem; it is distinct from the \(1/m\) uniform
attribution baseline over \(m\) input groups. Conformer under SBFP (3.85\%)
and Wav2Vec~2.0 in both panels (3.68\%) are near this chance-level regime. In
these rows, SA is \(0.100\), the uniform share over the ten audio channels,
whereas GB reaches \(0.206\) and \(0.198\) for Wav2Vec~2.0. The latter values
should be interpreted as attribution bias in the absence of a learned
predictive signal, rather than as superior recovery of the informative channel.

\section{Discussion and Conclusion}
\label{sec:discussion}

\paragraph{Attribution quality is model dependent.}
Across modalities, attribution should not be interpreted independently of the
underlying predictor. In most symbolic sweeps, informative-set identification
co-varies with the model's NPS, while relative ordering is largely stable over
the observed range. Near-chance models or poor symbolic approximations cannot
support strong identification claims. Yet good prediction is not sufficient
because a model may use irrelevant inputs and still predict accurately.
Prediction and attribution metrics must therefore be reported together.

\paragraph{Method complexity is not a guarantee.}
No method dominates all architectures and conditions. Simple input gradients
are frequently competitive and inexpensive, while perturbation, propagation,
surrogate, and path-based methods are stronger in particular settings. Method
choice should therefore reflect the predictor, modality, evaluation target,
and computational budget rather than presumed algorithmic sophistication.

\paragraph{Practical implications.}
Practitioners should pair held-out prediction with appropriate ground-truth
attribution metrics and separately report stability and computational cost,
using instance-specific localization when informative positions vary.

\paragraph{Conclusion.}
\name provides ground-truth symbolic, vision, and audio tasks for post-hoc
attribution with sparse informative inputs. Identification depends jointly on
predictive generalization, model and data conditions, and the attribution
procedure; additional complexity does not consistently improve it. Separately
reporting prediction, informative-set identification, stability, and cost
supports more qualified cross-modal comparisons.

\mainpageextension
\paragraph{Ethical statement.}
The benchmark collects no new human-subject data and combines publicly released
CIFAR-10, Oxford Flowers 102, Speech Commands, and Rainforest Connection audio
\cite{krizhevsky2009learning,Nilsback08,warden2018speech,
rfcx-species-audio-detection}. Reproduction materials are available at
\projectrepo{the project repository} under CC BY-NC 4.0.

\clearpage
\bibliographystyle{assets/plainnat}
\bibliography{paper}

\clearpage
\appendix
\section{Attribution Method Definitions}
\label{sec:appendix_background}

\subsection{Post-hoc Local Attribution Methods}
\label{subsec:posthoc_attribution}

Methods requiring a single baseline use the mean-feature reference in
Table~\ref{tab:attribution-settings}; DeepLIFT SHAP uses 50 reference samples.

\paragraph{Saliency (SA)}
Saliency uses the input gradient \cite{simonyan2013deep}:
\begin{equation}
    a_i=\frac{\partial F(x)}{\partial x_i}.
    \label{eq:sa}
\end{equation}

\paragraph{Integrated Gradients (IG)}
IG integrates gradients along the path from \(\bar{x}\) to \(x\)
\cite{sundararajan2017axiomatic}:
\begin{equation}
    a_i=(x_i-\bar{x}_i)\int_0^1
    \frac{\partial F(\bar{x}+\alpha(x-\bar{x}))}{\partial x_i}\,d\alpha.
    \label{eq:ig}
\end{equation}

\paragraph{DeepLIFT (DL)}
DeepLIFT propagates the reference-relative output difference using its rescale
rule \cite{shrikumar2017learning}:
\begin{equation}
\begin{aligned}
    r_i^{(l)}
    &=\sum_j
      \frac{z_{ji}-\bar{z}_{ji}}
      {\sum_{i'}z_{ji'}-\sum_{i'}\bar{z}_{ji'}}
      r_j^{(l+1)}, \\
    a_i&=r_i^{(0)},\qquad r^{(L)}=F(x)-F(\bar{x}).
\end{aligned}
    \label{eq:dl}
\end{equation}

\paragraph{Feature Ablation (FA)}
Feature Ablation measures the effect of replacing one coordinate by its
reference value \cite{zeiler2014visualizing}:
\begin{equation}
    a_i=F(x)-F(x_{[x_i=\bar{x}_i]}).
    \label{eq:fa}
\end{equation}

\paragraph{Kernel SHAP (KS)}
Kernel SHAP fits a locally weighted linear surrogate over coalitions of present
and absent features \cite{lundberg2017unified}:
\begin{equation}
\begin{aligned}
    (\phi_0,\phi)
    &=\arg\min_{\phi_0,\phi}
      \sum_{z'\in\{0,1\}^m}\pi_x(z') \\
    &\quad\times
      \left[F(h_x(z'))-\phi_0-\sum_{i=1}^{m}\phi_i z'_i\right]^2, \\
    a_i&=\phi_i.
\end{aligned}
    \label{eq:kernel_shap}
\end{equation}
Here \(h_x(z')\) replaces absent features by their reference values and
\(\pi_x\) is the SHAP coalition kernel.

\paragraph{LIME}
LIME fits a locally weighted interpretable surrogate
\cite{ribeiro2016why}:
\begin{equation}
\begin{aligned}
    g_x
    &=\arg\min_{g\in\mathcal{G}}
      \left\{\sum_{z'}\pi_x(z')
      \bigl(F(h_x(z'))-g(z')\bigr)^2+\Omega(g)\right\}, \\
    a_i&=\beta_i,\qquad
    g(z')=\beta_0+\sum_i\beta_i z'_i.
\end{aligned}
    \label{eq:lime}
\end{equation}

\paragraph{DeepLIFT SHAP (DLS)}
DeepLIFT SHAP averages DeepLIFT attributions over \(B\) reference inputs
\cite{shrikumar2017learning,lundberg2017unified}:
\begin{equation}
    a_i=\frac{1}{B}\sum_{b=1}^{B}
    A^{\mathrm{DL}}_i\!\left(x,\bar{x}^{(b)}\right).
    \label{eq:deeplift_shap}
\end{equation}

\paragraph{Guided Backpropagation (GB)}
Guided Backpropagation retains only positive upstream gradients through
positively activated ReLU units \cite{springenberg2015striving}:
\begin{equation}
    R_i^{(l)}
    =\mathbb{1}\!\left(x_i^{(l)}>0\right)
     \mathbb{1}\!\left(R_i^{(l+1)}>0\right)R_i^{(l+1)},
    \qquad a_i=R_i^{(0)}.
    \label{eq:guided_backprop}
\end{equation}

\section{Symbolic Function Catalog}
\label{sec:appendix_symbolic_functions}

\begin{table}[!t]
  \centering
  \caption{Symbolic generating functions and informative-set sizes. Variables
  appearing in a formula are informative; all additional coordinates are
  task-irrelevant.}
  \label{tab:symbolic}
  \scriptsize
  \setlength{\tabcolsep}{2.5pt}
  \renewcommand{\arraystretch}{1.08}
  \begin{tabularx}{\columnwidth}{@{}cc>{\raggedright\arraybackslash}X@{}}
    \toprule
    ID & $|S^{\star}|$ & Generating function $f_j$ \\
    \midrule
    1  & 1  & \(a\) \\
    2  & 1  & \(a^2\) \\
    3  & 1  & \(\dfrac{2}{a^2+1}-1\) \\
    4  & 1  & \(\sin(a)\) \\
    5  & 1  & \(\exp(a)-1.5\) \\
    6  & 1  & \(2\log(a^2+1)-1\) \\
    7  & 2  & \(0.25a^3+0.75b^2\) \\
    8  & 3  & \(0.5a^3+0.75b^2+ac\) \\
    9  & 4  & \(0.5e^a\sin(b)-0.25\dfrac{\cos^5(d)}{c^2+1}\) \\
    10 & 5  & \(0.5(a-b)^2+0.2(c+de)^3-0.5\) \\
    11 & 6  & \(\begin{aligned}
      &0.5\cos(a)\tan(b)\\[-2pt]
      &\quad-\dfrac{\log((c-d)^2+1)}{(e+f+1)^2+1}
    \end{aligned}\) \\
    12 & 7  & \(\begin{aligned}
      &0.5\dfrac{(b-c)^2}{a^2+1}+\tan(d)\log(e^2+1)\\[-2pt]
      &\quad+0.5\cos(f)\sin(g)
    \end{aligned}\) \\
    13 & 8  & \(\begin{aligned}
      &a+\tfrac12b^2+\tfrac13c^3+\tfrac14d^4\\[-2pt]
      &\quad+\tfrac15e^5+\tfrac16f^6+\tfrac17g^7+\tfrac18h^8
    \end{aligned}\) \\
    14 & 9  & \(\begin{aligned}
      &0.5\dfrac{a-1}{b^2+1}-0.5\dfrac{c^3}{d^2+1}
       +0.5\dfrac{e^5}{f^2+1}\\[-2pt]
      &\quad-0.5\dfrac{g^7}{h^2+1}+0.5\tan(i)+0.5
    \end{aligned}\) \\
    15 & 10 & \(\begin{aligned}
      &0.5\sin(a)-0.5b^3-\log(c^2+1)\\[-2pt]
      &\quad+0.5\sqrt{(d+e)^2+1}-0.5\cos(f)g\\[-2pt]
      &\quad+h\dfrac{i^2}{(1-j)^2+1}-0.5
    \end{aligned}\) \\
    \bottomrule
  \end{tabularx}
\end{table}

\end{document}